\documentclass[10pt,twocolumn,letterpaper]{article}

\usepackage{cvpr}              %

\usepackage{makecell}
\usepackage{tabularx}
\usepackage{color}
\usepackage{graphicx}
\usepackage{amssymb}%
\usepackage{pifont}%
\usepackage{bm}
\usepackage{multirow}
\usepackage{comment}
\usepackage{arydshln}
\usepackage[table]{xcolor}
\usepackage{cuted}
\usepackage{mathcomp}
\usepackage{siunitx}

\definecolor{cvprblue}{rgb}{0.21,0.49,0.74}
\usepackage[pagebackref,breaklinks,colorlinks,allcolors=cvprblue]{hyperref}

\newcommand{\cmark}{\textcolor{green!60!black}{\ding{51}}}
\newcommand{\xmark}{\textcolor{red!70!black}{\ding{55}}}

\def\paperID{924} %
\def\confName{CVPR}
\def\confYear{2026}

\title{Ghost-FWL: A Large-Scale Full-Waveform LiDAR Dataset \\ for Ghost Detection and Removal}

\author{
Kazuma Ikeda$^{1*}$,
Ryosei Hara$^{1*}$,
Rokuto Nagata$^{1}$,
Ozora Sako$^{1}$,
Zihao Ding$^{1}$, \\
Takahiro Kado$^{2}$,
Ibuki Fujioka$^{2}$,
Taro Beppu$^{2}$,
Mariko Isogawa$^{1}$,
Kentaro Yoshioka$^{1}$
\\[0.5em]
$^{1}$Keio University \hspace{2em} $^{2}$Sony Semiconductor Solutions
\hspace{2em} $^{*}$Equal contribution
}

\begin{document}

\twocolumn[{
\vspace{-8mm}
\maketitle
\vspace{-8mm}
\begin{center}
\includegraphics[width=1.0\linewidth]{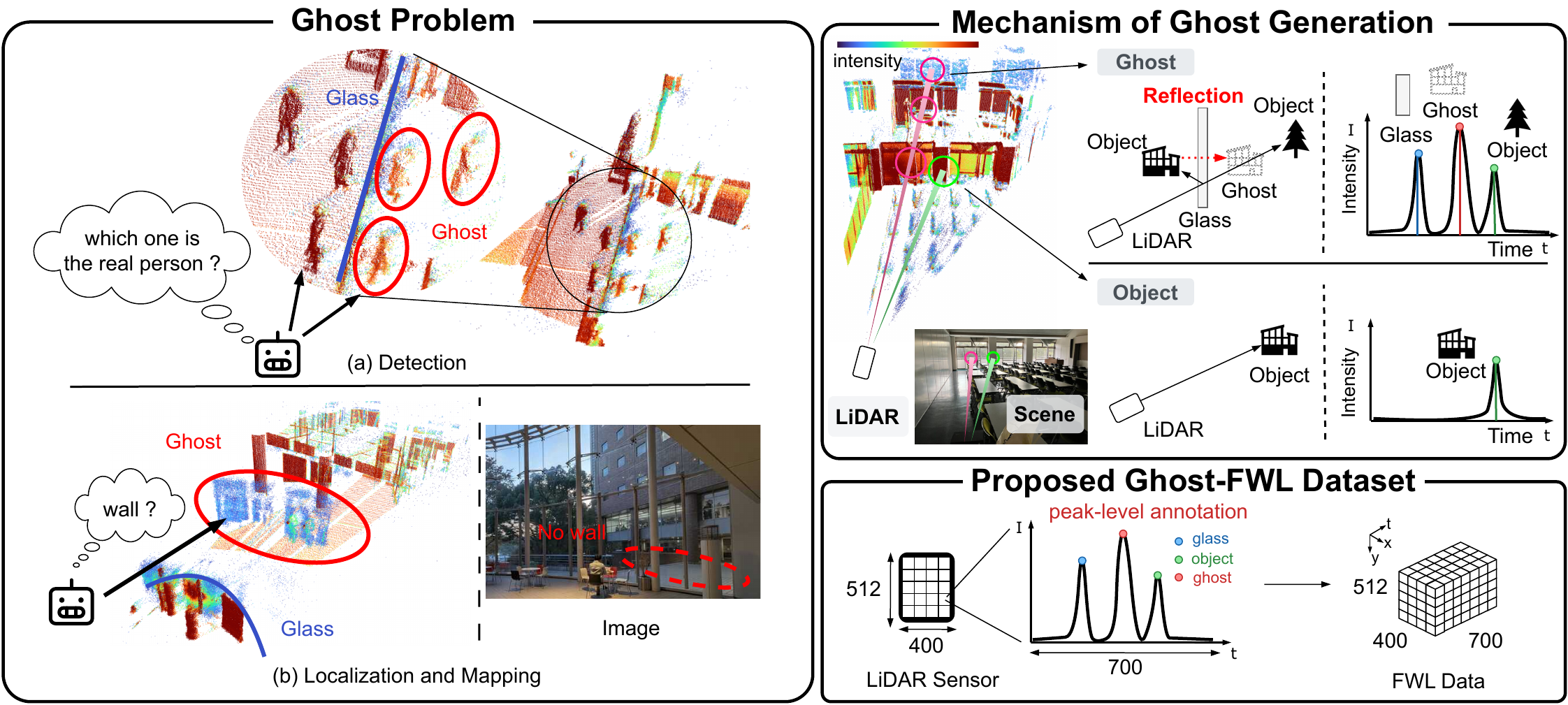}
\captionof{figure}{
    LiDAR data often contains \textbf{ghost} points caused by multi-path reflections from glass and reflective materials (top right), which appear as spurious structures that do not physically exist. Ghost leads to substantial errors in tasks such as detection (left (a)) and localization and mapping (left (b)). We address this issue by introducing the Ghost-FWL dataset (bottom right) and a ghost removal framework.
    }
\label{fig:teaser}
\end{center}
}]

\begin{abstract}
\vspace{-7mm}

LiDAR has become an essential sensing modality in autonomous driving, robotics, and smart-city applications.
However, ghost points (or \textbf{ghost}), which are false reflections caused by multi-path laser returns from glass and reflective surfaces, severely degrade 3D mapping and localization accuracy.
Prior ghost removal relies on geometric consistency in dense point clouds, failing on mobile LiDAR's sparse, dynamic data.
We address this by exploiting full-waveform LiDAR (FWL), which captures complete temporal intensity profiles rather than just peak distances, providing crucial cues for distinguishing ghosts from genuine reflections in mobile scenarios. As this is a new task, we present Ghost-FWL, the first and largest annotated mobile FWL dataset for ghost detection and removal. Ghost-FWL comprises 24K frames across 10 diverse scenes with 7.5 billion peak-level annotations, which is 100$\times$ larger than existing annotated FWL datasets.
Benefiting from this large-scale dataset, we establish a FWL-based baseline model for ghost detection and propose FWL-MAE, a masked autoencoder for efficient self-supervised representation learning on FWL data.
Experiments show that our baseline outperforms existing methods in ghost removal accuracy, and our ghost removal further enhances downstream tasks such as LiDAR-based SLAM (66\% trajectory error reduction) and 3D object detection (50$\times$ false positive reduction).
The dataset and code is publicly available and can be accessed via the project page: \url{https://keio-csg.github.io/Ghost-FWL/}.

\end{abstract}

\vspace{-4mm}

\section{Introduction}
\label{sec:intro}

LiDAR (Light Detection And Ranging) is a well-used range-sensor that measures the time of flight of emitted laser pulses reflected from surrounding objects, enabling 3D geometry reconstruction of the scene. Given its long-range sensing capability and robustness to illumination changes, LiDAR has become an indispensable sensor in a wide range of applications such as autonomous driving~\cite{autoware, apollo}, robotics~\cite{unitreeg1, spot}, and large-scale terrain mapping~\cite{airborne-lidar}.

However, LiDAR often suffers from a critical issue of false detections, commonly referred to as ``\textbf{ghost}''. Ghosts occur when emitted laser pulses are reflected by transparent or reflective surfaces such as glass, causing spurious LiDAR 3D points to appear at non-existent locations (~\cref{fig:teaser}).
This challenge has grown with recent LiDAR advancements: increased sensor sensitivity improves detection range but simultaneously amplifies weak multi-path returns, making ghosts more prevalent in modern systems~\cite{at128, m3}.
These artifacts can lead to severe failures in downstream tasks, such as producing false positives in object detection (~\cref{fig:teaser} (a)) or generating incorrect 3D maps and causing localization collapse in SLAM (~\cref{fig:teaser} (b)). %

Prior works~\cite{yun2018reflection, yun2019virtual, zhao2020mapping, lee2023learning} have attempted to remove ghosts by leveraging geometric consistency between points. 
However, these methods assume static, high-density scanning setups used in construction or terrain surveying and do not generalize to mobile LiDAR systems with sparse point clouds.
In real-world robotics and autonomous driving scenarios, where LiDARs must operate in dynamic and reflective environments, ghost removal remains unsolved due to the limited geometric cues available per frame.

To address this limitation, full-waveform LiDAR (FWL) offers a promising alternative. Unlike point-based measurements that record only peak distances, FWL captures the complete temporal intensity profile of each laser pulse, encoding both direct and indirect returns.
This richer signal provides intensity and temporal cues that could enable more robust ghost detection in mobile scenarios.
However, no dataset exists to enable learning-based ghost removal from FWL data. Existing LiDAR datasets~\cite{caesar2020nuscenes, sun2020waymo, geiger2013kitti} focus on point clouds and do not include full-waveform data. While a few ghost detection datasets exist~\cite{yun2019virtual, lee2023learning}, they rely on stationary, high-precision scanners unsuitable for mobile systems. The only public FWL dataset, PixSet~\cite{deziel2021pixset}, lacks peak-level annotations necessary to distinguish ghost peaks from genuine reflections and does not address ghost phenomena. Moreover, reproducing ghosts in simulation requires modeling multi-path reflections, which is computationally expensive and physically inaccurate~\cite{scheublelidar}, making synthetic data generation impractical.

Therefore, we present \textbf{Ghost-FWL}, the first full-waveform LiDAR dataset for ghost detection and removal in mobile scenarios. Ghost-FWL contains 24K annotated frames collected across 10 diverse indoor and outdoor scenes, providing 7.5 billion peak-level labels for ghost, glass, object, and noise reflections. With complete temporal intensity profiles captured from real-world mobile LiDAR, Ghost-FWL is 100$\times$ larger than prior work~\cite{scheublelidar} and is the largest annotated FWL dataset to date. Unlike previous datasets that rely on stationary scanners~\cite{yun2019virtual} or lack peak-level labels~\cite{deziel2021pixset}, Ghost-FWL reflects practical mobile conditions with sparse, dynamic data and diverse reflective environments—including building facades, glass storefronts, and interior surfaces under varying viewing angles and illumination.

We propose the first baseline to tackle ghost removal using FWL data since no prior work has addressed this task. To enable effective training despite the high annotation cost of peak-level labeling, we further introduce FWL-MAE, a masked autoencoder designed for FWL data. 
Unlike existing MAE approaches designed for images~\cite{he2022mae} or transient images~\cite{shen2025marmot}, FWL-MAE performs self-supervised pre-training on unlabeled data by reconstructing masked temporal regions while explicitly modeling peak properties (position, amplitude, and width) to learn representations that better capture the underlying physical characteristics of FWL data.

Experimental results show that our baseline with FWL-MAE outperforms other existing methods in terms of ghost detection accuracy.
Furthermore, when applied to downstream tasks such as SLAM and 3D object detection, our baseline significantly improves performance in ghost-existing environments, achieving up to 66\% trajectory error reduction and 50$\times$ reduced ghost-induced false-positives.

To summarize, our main contributions are as follows:
\begin{itemize}
    \item We present the \textbf{Ghost-FWL dataset}, the largest annotated mobile full-waveform LiDAR dataset, comprising 7.5B peak-level annotations across 24K frames and 10 diverse real-world scenarios, which is more than 100 times larger than previous datasets.
    \item We are the first to propose the FWL-based ghost-removal baseline method. To enable effective training, we further propose \textbf{FWL-MAE}, a masked autoencoder designed for FWL data.
    \item Experimental results show that our baseline with FWL-MAE outperforms existing methods and significantly improves downstream performance, enhancing LiDAR-based SLAM and 3D object detection in ghost-affected environments.
\end{itemize}

\section{Related Work}
\label{sec:related}

\begin{table*}[t]
\centering
\caption{
\textbf{Comparison of LiDAR real-world datasets for ghost detection and/or full-waveform analysis.} Our Ghost-FWL contains mobile LiDAR full-waveform measurements and is one hundred times larger than prior work, making it the largest annotated FWL dataset.
}
\vspace{-2mm}
\label{tab:fwldataset_comparison}
\setlength{\tabcolsep}{2pt}
\renewcommand{\arraystretch}{0.98}
\begin{tabularx}{\textwidth}{l*{10}{>{\centering\arraybackslash}X}}
\toprule
& \multicolumn{3}{c}{\textbf{Access \& Platform}}
& \multicolumn{3}{c}{\textbf{Sensor}}
& \multicolumn{4}{c}{\textbf{Labels}} \\
\cmidrule(lr){2-4} \cmidrule(lr){5-7} \cmidrule(lr){8-11}
\textbf{Dataset} & Year & Public & Platform & FWL & LiDAR Dim. & Ray Den. & Ghost & FWL Data & Frames/ Scenes\textsuperscript{†} & Annotated Peaks \\
\midrule
UNIST~\cite{yun2019virtual} & 2017 & \cmark & Stationary & \xmark & 3D & 278 & \cmark & \xmark & -- & -- \\
Leddar PixSet~\cite{deziel2021pixset} & 2021 & \cmark & Mobile & \textbf{\cmark} & 3D & 0.267 & \xmark & \cmark & -- & -- \\
Lee et al.~\cite{lee2023learning} & 2023 & \xmark & Stationary & \xmark & 3D & 278 & \xmark & \xmark & -- & -- \\
FRACTAL~\cite{gaydon2024fractal} & 2024 & \cmark & Aerial & \xmark & 2D & -- & \xmark & \xmark & -- & -- \\
Scheuble et al.~\cite{scheublelidar} & 2025 & \xmark & Mobile & \cmark & 3D & 2.56 & \xmark & \cmark & 0.24k / 2 & NA \\
\midrule\midrule
\textbf{Ghost-FWL~(Ours)} & \textbf{2025} & \cmark & \textbf{Mobile} & \textbf{\cmark} & \textbf{3D} & \textbf{200} & \textbf{\cmark} & \textbf{\cmark} & \textbf{24k / 10} & \textbf{7.5B} \\
\bottomrule
\end{tabularx}
\vspace{2pt}
\footnotesize
FWL: Full-Waveform LiDAR. \textsuperscript{†}Frames/Scenes: number of annotated frames and number of scenes within the real-world FWL data.
\end{table*}

\subsection{Ghost Point Detection and Removal}
Ghost points arise from multi-path reflections off transparent or reflective surfaces, degrading 3D reconstruction and localization. Prior methods address this through geometric consistency. Optimization-based approaches \cite{yun2018reflection, yun2019virtual, zhao2020mapping} exploit symmetry properties or statistical features to identify ghosts, but struggle with noise, complex structures, and multiple reflective surfaces. Learning-based methods~\cite{lee2023learning} combine geometric features with deep networks, yet remain limited to static, high-density scans where geometric cues are abundant. These approaches fail in mobile scenarios with sparse, single-frame data typical of robots and autonomous driving, where geometric consistency cannot be reliably established. In contrast, our work leverages FWL data that encode temporal and intensity information beyond geometric cues, enabling ghost detection in challenging mobile environments.

\subsection{FWL Processing on Mobile LiDAR Platforms}

While conventional LiDAR records only distance information, FWL captures the complete temporal intensity profile of each laser pulse~\cite{mallet2009full}. By leveraging peak characteristics contained in this rich waveform information, numerous studies have been conducted to improve ranging accuracy and measurement reliability in mobile LiDAR~\cite{reynolds2011capturing}.
Early approaches primarily integrated waveform information through rule-based or CNN-based feature extraction~\cite{tanabe2019inter, yoshioka201820, zou2024256} to improve ranging accuracy. However, all these methods rely on rule-based peak detection and fail to fully exploit the spatial and temporal correlations inherent in FWL data.
Scheuble et al.~\cite{scheublelidar} proposed an end-to-end learning framework that takes entire FWL data as input and jointly learns peak detection and range estimation, improving ranging accuracy and denoising performance under foggy conditions. This data-driven peak detection approach effectively leverages spatial and temporal features across the entire waveform.

Our method similarly leverages end-to-end learning on complete FWL data. The key difference lies in the task formulation: whereas prior works target range estimation, we directly learn temporal peak structures to classify peak origins based on their physical causes: Object, Glass, or Ghost. In doing so, we broaden the scope of FWL processing from its conventional ``ranging-centric'' focus to encompass the physical interpretation of FWL data.

\subsection{LiDAR Datasets and Full-Waveform Data}
Large-scale LiDAR datasets such as nuScenes~\cite{caesar2020nuscenes}, Waymo~\cite{sun2020waymo}, and KITTI~\cite{geiger2013kitti} have driven progress in 3D object detection and autonomous driving. However, these datasets provide only conventional point clouds without FWL data or ghost annotations.
A few datasets target ghost detection, notably UNIST LS3DPC~\cite{yun2019virtual}, but rely on stationary, high-precision scanners in controlled settings unsuitable for mobile platforms. 

FWL captures complete temporal intensity profiles rather than single-peak distances, recording multi-path returns crucial for ghost detection. PixSet~\cite{deziel2021pixset} is the only public FWL dataset, yet it lacks peak-level annotations and does not address ghost phenomena. Scheuble et al.~\cite{scheublelidar} also utilized mobile FWL data and applied machine learning to improve range estimation. However, their study focused on enhancing measurement accuracy rather than ghost detection, and the dataset itself was not released publicly. Table~\ref{tab:fwldataset_comparison} compares existing datasets, revealing a critical gap: no publicly available mobile FWL dataset exists with ghost-specific peak-level annotations.

An alternative approach would be to synthesize such data. However, reproducing ghosts in simulation requires modeling multi-path reflections, which is computationally expensive and physically inaccurate; CARLA~\cite{dosovitskiy2017carla} lacks multi-bounce support and Mitsuba~\cite{Mitsuba3} requires extensive tuning for outdoor scenes. 
We address these limitations by constructing Ghost-FWL, a large-scale real-world FWL dataset with peak-level annotations for ghost, glass, and object reflections, without reliance on synthetic data.

\subsection{Representation Learning with Self-Supervision}
\label{sec2:pretrain}

\begin{figure*}[tp]
  \begin{center}
    \includegraphics[width=1.0\linewidth]{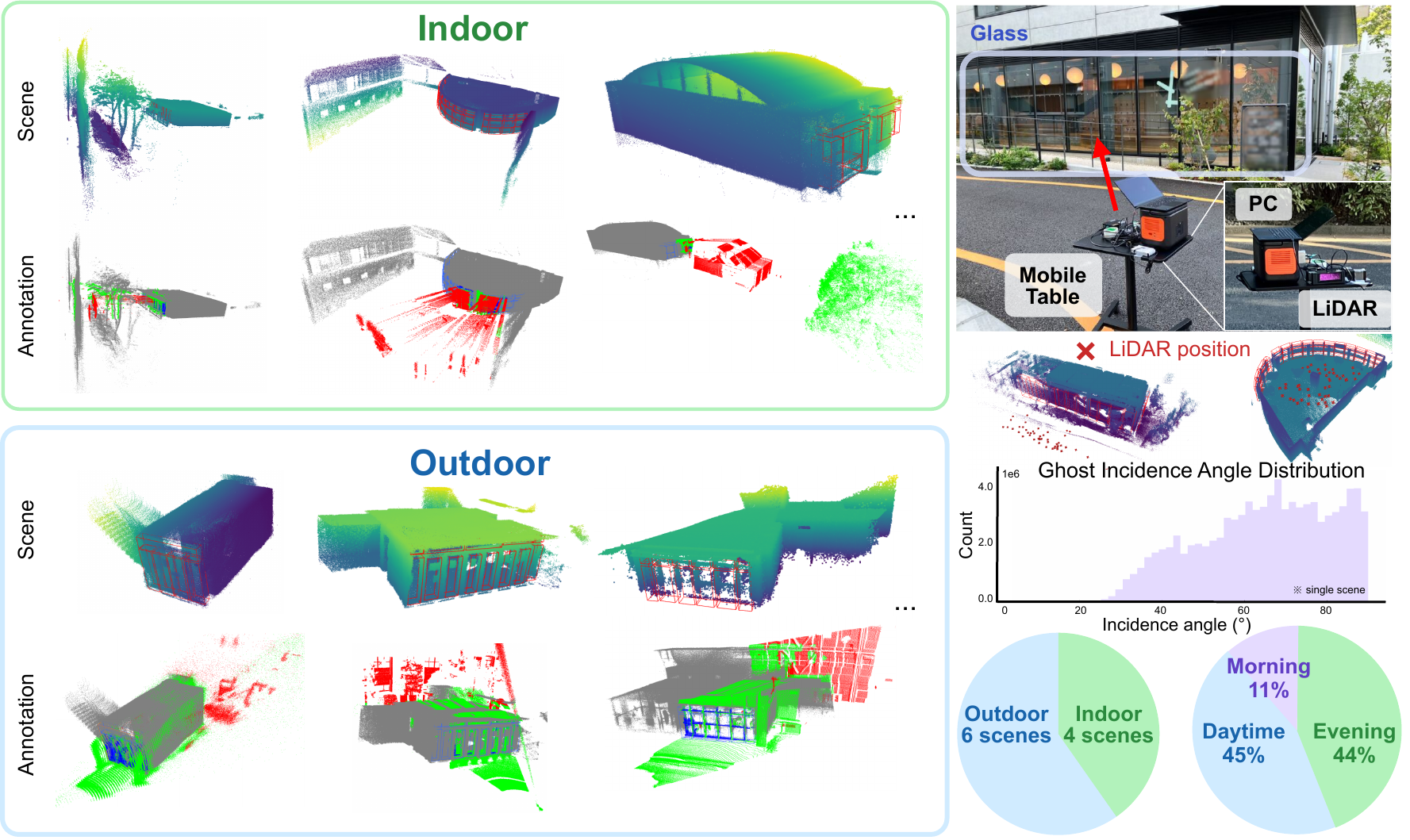}
    \vspace{-4mm}
    \caption{\textbf{Overview of the Ghost-FWL.} 
    Left: Our dataset includes both indoor and outdoor scenes. Based on the dense 3D maps as shown in Scene, we annotated FWL data with semantic labels: \textit{Ghost} (red), \textit{Object} (green), \textit{Glass} (blue), \textit{Noise}. Gray regions are excluded from annotation. Right shows the data acquisition setup and dataset statistics, including the incident angle distribution and LiDAR positions examples. Data were collected at three different times of day: Morning (AM10–12), Daytime (PM12–5), and Evening (PM5–7).
    }
    \label{fig:dataset_scene}
  \end{center}
  \vspace{-6mm}
\end{figure*}

Acquiring informative data representations is important for effective model training. However, obtaining large-scale annotated datasets is costly, which motivates research on self-supervised learning that reduces reliance on manual labels. Contrastive methods such as SimCLR~\cite{chen2020simclr} and MoCo~\cite{he2020moco, chen2020moco2, chen2021moco3} learn generalizable features by aligning views of the same instance while separating different ones. Masked Autoencoders (MAE)~\cite{he2022mae} further advance this paradigm by reconstructing masked regions from visible inputs, enabling the model to capture structural regularities in an unsupervised manner. This idea has been extended to videos~\cite{tong2022videomae}, 3D point clouds~\cite{pang2022pointmae, liu2022maskpoint, yu2022pointbert}, and voxel data~\cite{hess2023voxelmae, xu2023mvjar, yang2023gdmae, tian2023geomae, min2024occupancymae}.

The work most relevant to ours is MARMOT~\cite{shen2025marmot} which focuses on temporal histogram data. 
MARMOT learns representations of transient images containing spatiotemporal 3D information by randomly masking and reconstructing parts of the input. However, it primarily performs voxel-level reconstruction and does not explicitly account for histogram-specific statistical properties such as intensity peak locations or distribution shapes. In contrast, we propose a representation learning model specialized for histogram data called FWL-MAE that explicitly models temporal continuity and peak information in FWL data.

\begin{figure*}[tb]
  \begin{center}
    \includegraphics[width=0.9\textwidth]{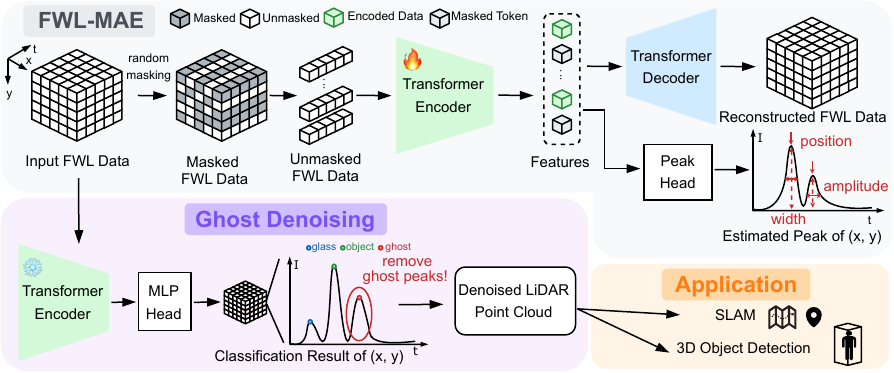}
    \vspace{-2mm}
    \caption{\textbf{FWL-based ghost removal framework.}
    Given FWL data, our framework predicts and removes ghost-related signals. Our model consists of a transformer-based encoder and an MLP head. We further introduce FWL-MAE, a masked autoencoder designed for representation learning on FWL data, explicitly trained to reconstruct peak \textit{position}, \textit{amplitude} and \textit{width}. The ghosts detected by our model are then removed from FWL data, and the cleaned data are utilized for downstream tasks such as SLAM and 3D object detection.    
    }
    \label{fig:framework}
  \end{center}
  \vspace{-6mm}
\end{figure*}

\section{Ghost-FWL Dataset}
This section presents Ghost-FWL, the largest FWL dataset to date, which is specialized for ghost removal. Conventional LiDAR datasets provide only point cloud-level information, discarding the temporal multi-path information crucial for identifying ghosts caused by glass and reflective surfaces. Ghost-FWL addresses this gap by capturing complete temporal intensity histograms and providing peak-level annotations indicating the physical cause of each reflection (object, glass, ghost, or noise). Spanning 10 diverse scenes with 24,412 annotated frames and 7.5B peak-level labels, Ghost-FWL is 100$\times$ larger than prior annotated FWL datasets~\cite{scheublelidar}, enabling learning-based ghost detection and removal at the waveform level. Statistics of the dataset are shown in ~\cref{fig:dataset_scene}.

\subsection{Sensing System and Data Collection}
\noindent\textbf{Custom FWL Acquisition System:}
Commercial LiDAR devices typically output only processed 3D point clouds containing range and intensity information, without providing access to the underlying full-waveform data. To overcome this limitation, we developed a custom acquisition system that directly accesses the FPGA module of the LiDAR hardware, extracting raw FWL data from the internal signal processing pipeline. This enables frame-by-frame capture of the complete received signal, preserving reflection peaks and multi-path components essential for ghost detection.

\noindent\textbf{Sensor Specifications:} 

The FWL sensor system produces histograms of 512 × 400 pixels (vertical × horizontal), recording up to 700 temporal bins per ranging direction with approximately 1 ns time resolution (max. range 105 m). %

\subsection{Sensing Scenario and Scene}

To capture diverse ghosts in real-world conditions, we collected data across 10 scenes (4 indoor, 6 outdoor) totaling 24,412 frames. The FWL sensor was mounted on a mobile platform to simulate mobile LiDAR scenarios common in robotics and autonomous driving.

\noindent\textbf{Scene Selection:}
Indoor scenes include office floors, communication lounges, and gymnasiums; spaces featuring large glass walls where lighting and surface reflections create complex multi-path conditions. Outdoor scenes comprise building entrances, glass-curtain facades, and glass-lined pedestrian areas, providing natural lighting variations, changing incident angles, and long-range reflections characteristic of autonomous driving environments.

\noindent\textbf{Environmental Diversity:}
We systematically varied environmental conditions to ensure dataset diversity. Data collection spanned different times of day (morning to evening) to capture varying illumination effects on waveform characteristics. Within each scene, we varied sensor-to-glass distance (3-20 m) and incident angle (0$\tcdegree$–40$\tcdegree$) to comprehensively capture reflection behavior under different geometric conditions. Beyond static environments, selected scenes include dynamic elements such as pedestrians and moving objects, reflecting realistic robotics and autonomous driving conditions.

\noindent\textbf{Data Collection Protocol:}

We employed two capture strategies serving different learning objectives, collecting 33,345 total frames:
\textbf{(1) Multi-Viewpoint Static Capture:} 
At each scene, we selected 37–55 viewpoints and systematically varied the incident angle on glass surfaces and sensor orientation at each location. Capturing approximately 50 frames per viewpoint yielded an average of 2,441 frames per scene, totaling 24,412 annotated frames for supervised ghost detection.
\noindent\textbf{(2) Mobile Trajectory Capture:}
We recorded continuous mobile trajectories through each scene (500–1,500 frames per scene, 8,933 total), simulating realistic robotic operation. These sequences remain unlabeled, as continuous motion makes peak-level annotation prohibitively expensive, but provide diverse data for self-supervised pre-training in \S\ref{sec:method}.

\subsection{Annotation}

As shown in ~\cref{fig:teaser}, we annotated FWL data at the peak level, assigning each reflection peak exceeding a threshold to one of four classes based on its physical origin: Object, Glass, Ghost, or Noise.
Annotation followed a semi-automatic pipeline leveraging high-precision 3D map point clouds generated via SLAM. First, we constructed a 3D map of each scene using a commercial 360$\tcdegree$ LiDAR sensor (Livox Mid-360~\cite{mid-360}), removed noise, and manually annotated glass surface regions and solid object regions. Next, we converted peak positions extracted from FWL data into point clouds and performed coordinate alignment with the 3D map. This established correspondence between each reflection peak and real-world scene structures.

Peaks were then automatically classified according to the following criteria: (1)~\textit{Object}: peaks generating FWL-derived points in close proximity to the 3D map. (2)~\textit{Glass}: peaks exhibiting surface reflections within annotated glass regions. (3)~\textit{Ghost}: peaks appearing at locations not corresponding to the 3D map after passing through or reflecting off glass. (4)~\textit{Noise}: remaining noise or weak reflections. Finally, annotations were reviewed by domain experts with expertise in computer vision and LiDAR sensing.

\section{FWL-based Ghost Removal Framework}
\label{sec:method}
As we are the first to address this task, we propose a baseline framework that detects and removes ghost-related peaks directly from FWL data (~\cref{fig:framework}).
The framework first performs classification on FWL data $\bm{V} \in \mathbb{R}^{H \times W \times T}$ to identify ghost-related regions, then removes the corresponding 3D points predicted as Ghost.
To obtain more discriminative representations from limited FWL data, we further propose FWL-MAE, a Masked Autoencoder tailored for FWL data that models their inherent peak structures.

\subsection{Full Waveform LiDAR Masked Autoencoder}
We propose the Full-Waveform LiDAR Masked Autoencoder (FWL-MAE), a self-supervised pretraining method specifically designed to learn latent representations from FWL data. Our approach is inspired by the Masked Autoencoder (MAE)~\cite{he2022mae}, which learns data representations from 2D images, and by MARMOT~\cite{shen2025marmot}, which extends this concept to transient histograms.

Following MARMOT, FWL-MAE takes FWL data $\bm{V} \in \mathbb{R}^{H \times W \times T}$ as input, randomly samples spatial patches in the $(x, y)$ region, and masks all temporal bins along the $T$ axis within each selected patch.
FWL-MAE trains a Transformer-based encoder that outputs a latent representation $f_{\theta}(\bm{V})$ from an input histogram volume $\bm{V}$, where $\theta$ is a trainable parameter. The encoder consists of six Transformer blocks with six attention heads in each block.
Unlike MARMOT, FWL-MAE additionally estimates the position, amplitude and width of histogram peaks using a linear head to capture physically meaningful latent representations from real FWL data. 

\noindent\textbf{Loss Function.}
Following the original MAE~\cite{he2022mae}, we use the mean squared error loss ($\mathcal{L}_{\text{MSE}}$) to evaluate the reconstruction accuracy in the voxel region.
To assess the distance between predicted and ground-truth values of the dominant peak \textit{position} ($p$), \textit{amplitude} ($a$), and \textit{width} ($w$) within each waveform, we employ an L1 loss, denoted as $\mathcal{L}_1^{\text{peak-}p}$, $\mathcal{L}_1^{\text{peak-}a}$, and $\mathcal{L}_1^{\text{peak-}w}$ for each attribute, respectively.
The overall loss of FWL-MAE, $\mathcal{L}_{\text{FWL-MAE}}$, is defined as the weighted sum of these losses as follows:
\begin{equation}
\mathcal{L}_{\text{FWL-MAE}} =
\mathcal{L_{\text{MSE}}}
+ \lambda_p \mathcal{L}_1^{\text{peak-}p}
+ \lambda_a \mathcal{L}_1^{\text{peak-}a}
+ \lambda_w \mathcal{L}_1^{\text{peak-}w},
\end{equation}
where $\lambda_p$, $\lambda_a$, and $\lambda_w$ are hyperparameters that control the contribution of each term.

\subsection{Ghost Detection and Removal} 
To detect and remove ghosts, our method takes the FWL data $\bm{V}$ as input and estimates the class probabilities $\bm{P} \in \mathbb{R}^{H \times W \times T \times C}$ for the categories \textit{Glass}, \textit{Ghost}, \textit{Object}, and \textit{Noise}.
To extract informative features from the FWL data, we use the encoder pretrained with FWL-MAE and keep its weights frozen to obtain latent representations $f_{\theta}(\bm{V})$.
A lightweight classification head composed of two linear layers is then applied to predict the class probabilities for all FWL data coordinates.
By removing the 3D points corresponding to histogram peaks predicted as Ghost, we obtain a filtered LiDAR point cloud.
The denoised point cloud produced by this framework can subsequently be used as input for downstream tasks such as SLAM or 3D object detection.

\noindent\textbf{Loss Function.}
We adopt the focal loss~\cite{lin2017focal}, which mitigates the impact of class imbalance in multi-class classification.
Our task poses a challenging classification problem
involving highly imbalanced data, where minority classes such as \textit{Ghost} coexist with the majority \textit{Noise} class.

\section{Experiments and Results}
This section comprehensively evaluated the effectiveness of our proposed ghost removal framework. We first conducted a quantitative evaluation of its ability to classify and remove ghost points in \ref{subsec:ghost-denoising-evaluation}, and then \ref{subsec:downstream-evaluation} investigated how the denoised data affects the performance of downstream tasks, including SLAM and 3D object detection. 
Implementation details and hyperparameters are provided in the supplementary material.

\begin{figure}[tp]
  \begin{center}
    \includegraphics[width=1.0\linewidth]{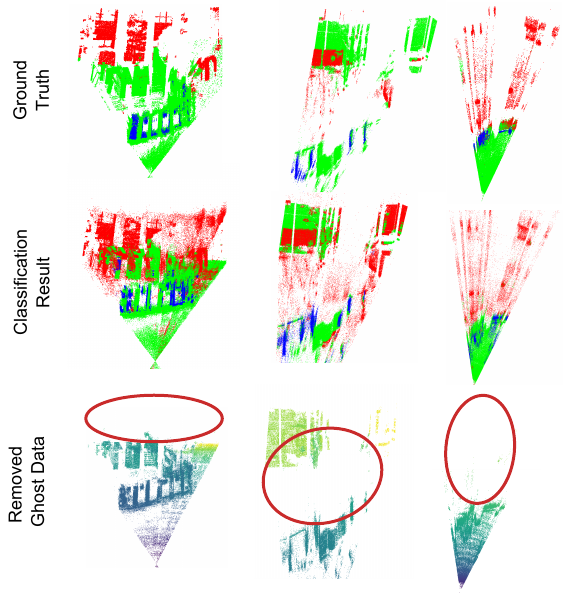}
    \vspace{-6mm}
    \caption{Peak classification results and point cloud visualization after applying ghost removal. All results were obtained using the proposed framework. Red, green and blue indicates \textit{Ghost}, \textit{Object}, \textit{Glass}, respectively.}
    \label{fig:removal_result}
  \end{center}
  \vspace{-12mm}
\end{figure}

\subsection{Ghost Denoising Evaluation}
\label{subsec:ghost-denoising-evaluation}
This subsection evaluated the performance of our FWL-MAE framework for ghost denoising. The task is formulated as a four-class classification problem with $C = 4$ (Glass, Ghost, Object, and Noise).

\noindent\textbf{Experimental Settings.~}
All models were trained using our Ghost-FWL dataset.
The raw FWL data were reshaped to $(H, W, T) = (128, 128, 256)$ before being fed into the model through the following preprocessing. First, the top and bottom 90 bins corresponding to reflections from the ceiling and floor and the front 25 bins containing noise from internal sensor reflections were removed. 
The remaining data were then uniformly down-sampled along the $T$ axis and randomly cropped into $H \times W$ size.
We used a data split of 13,853 for training, 2,994 for validation, and 1,427 for testing. 

\noindent\textbf{Metrics.~}
We evaluate ghost detection at two levels: peak-level and point-level. For peak-level evaluation, we follow Scheuble et al.~\cite{scheublelidar} and report recall, measuring the proportion of correctly detected ghost peaks among all ground-truth ghost peaks. 
For point-level evaluation, we introduce the \textit{Ghost Removal Rate}, which measures the proportion of ghost points successfully removed after converting predicted peaks to 3D point clouds. This metric is inspired by the snow removal rate in Charron et al.~\cite{charron2018noising} and assesses practical denoising effectiveness in downstream tasks.

\noindent\textbf{Comparative Methods.~}
To investigate the effectiveness of the proposed FWL-MAE pretrained encoder, we compared three models: (1) the proposed model incorporating FWL-MAE (\textbf{Ours}), (2) the model without FWL-MAE (\textbf{Ours w/o FWL-MAE}), and (3) the model pretrained using a general MAE designed for transient imaging (\textbf{MARMOT}~\cite{shen2025marmot}).

\noindent\textbf{Results.~}
As shown in Table~\ref{tab:ghost_class_evaluation}, the proposed model with FWL-MAE achieves superior performance in both ghost detection recall and ghost removal rate compared to other models. These results demonstrate the effectiveness of incorporating FWL-MAE, which enables pretraining that better captures the physical characteristics of FWL data.

~\cref{fig:removal_result} shows the qualitative evaluation. 
It can be seen that the method successfully eliminates various types of ghost artifacts, including those spreading horizontally or vertically, and those occurring between buildings.
The proposed Ghost Removal Framework clearly achieves accurate class classification and effective ghost removal.

\begin{table}[t]
\centering
\caption{Comparison of ghost removal performance with other methods.}
\vspace{-2mm}
\label{tab:ghost_class_evaluation}
\setlength{\tabcolsep}{6pt}
\renewcommand{\arraystretch}{1.0}

\begin{tabularx}{0.9\linewidth}{l>{\centering\arraybackslash}X>
{\centering\arraybackslash}X
}
\toprule
Method &  Recall ($\uparrow$)  & Removal Rate ($\uparrow$) \\
\midrule
MARMOT~\cite{shen2025marmot} & 0.746 & 0.910 \\
Ours w/o FWL-MAE  & 0.704 & 0.900  \\
\midrule
Ours & \textbf{0.751} & \textbf{0.918}  \\
\bottomrule
\end{tabularx}
\vspace{-4mm}
\end{table}

\subsection{Evaluation on Downstream Applications}
\label{subsec:downstream-evaluation}

\subsubsection{Evaluation on SLAM}
\label{subsec:slam-evaluation}
The presence of ghost points introduces false geometric structures that do not exist in the real world, leading to frame misalignment and degraded SLAM performance. This problem is particularly critical in autonomous driving, where even small localization errors can have serious consequences. For example, a position error of only 0.29 m on regular roads or 0.74 m on highways can cause a vehicle to drift into a lane boundary~\cite{sato-dirty}. This experiment evaluates whether the proposed ghost removal framework can mitigate such issues in real-world SLAM scenarios.

\noindent\textbf{Experimental Settings.~}
We collected SLAM test sequences using the same sensor configuration as the Ghost-FWL dataset setup. The data was captured at three locations: an outdoor area of the office building and two indoor locations of the office building, including an indoor space and an indoor office corridor where strong reflections frequently produce ghost points. The sequence contains 231 frames along a 23.4 m trajectory. The ground-truth trajectory was obtained using multiple fixed high-precision LiDAR sensors that continuously tracked the position of the moving FWL sensor. Ghost removal followed the point-level evaluation pipeline from \S\ref{subsec:ghost-denoising-evaluation}.

\noindent\textbf{Metrics.~}
We evaluated SLAM performance using standard metrics: Absolute Trajectory Error (ATE), Relative Trajectory Error (RTE) \cite{zhang2018tutorial}. Following~\cite{koide2024glim}, we report both mean and standard deviation of ATE and RTE, where ATE measures the Euclidean distance between each estimated pose and its nearest ground-truth pose after alignment.

\noindent\textbf{Comparative Methods.~}
We compare against LiDAR signal processing strategies using their factory default settings: Dual-Peak~\cite{ouster,at128} and Multi-Peak~\cite{avia}, which retain the two and three strongest intensity peaks, respectively.
All methods employed the same SLAM backend, GLIM~\cite{koide2024glim} for fair comparison.

\begin{figure}[tp]
  \begin{center}
    \includegraphics[width=1.0\linewidth]{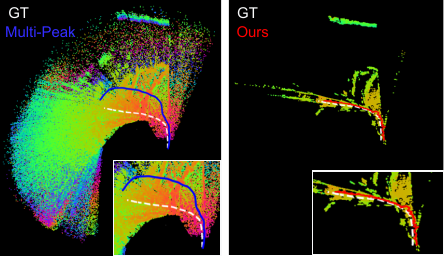}
    \vspace{-6mm}
    \caption{
    Trajectory and mapping generated by SLAM using Multi-Peak processing (left) and our ghost removal method (right). Multi-Peak processing includes numerous ghost points in the reconstructed map, leading to trajectory drift. The proposed method yields a trajectory that more closely follows the ground-truth path (white) by effectively removing ghost artifacts.    
    }
    \label{fig:slam_map_comparison}
  \end{center}
  \vspace{-6mm}
\end{figure}

\noindent\textbf{Results.}
Table~\ref{tab:slam_comparison} shows that our ghost removal significantly improved SLAM accuracy, reducing ATE by 66–84\% and RTE by 67–85\% compared to baseline peak selection methods. The improvement was most pronounced near glass surfaces, where cumulative localization errors from ghost reflections were effectively suppressed. 
\cref{fig:slam_map_comparison} illustrates successful ghost removal behind glass while maintaining accurate localization and consistent 3D map reconstruction.

\begin{table}[t]
\centering
\caption{SLAM performance under different LiDAR signal preprocessing methods.}
\label{tab:slam_comparison}
\setlength{\tabcolsep}{3pt} %
\renewcommand{\arraystretch}{0.95} %
\begin{tabularx}{0.48\textwidth}{l*{2}{>{\centering\arraybackslash}X}}
\toprule
Method & ATE~[m] ($\downarrow$) & RTE~[m] ($\downarrow$) \\
\midrule
Dual-Peak~\cite{ouster,at128}      & 0.715$\pm$0.433  & 0.741$\pm$0.406 \\
Multi-Peak~\cite{avia}     & 1.547$\pm$1.394 & 1.602$\pm$1.381 \\
\midrule
Ours                        & \textbf{0.245$\pm$0.138} & \textbf{0.245$\pm$0.131} \\
\bottomrule
\end{tabularx}
\vspace{-4mm}
\end{table}

\subsubsection{Evaluation on Object Detection}
Ghost points often cause serious safety risks with false positives in 3D object detection. In particular, mirrored ghosts of vehicles or pedestrians appearing behind glass surfaces can propagate through tracking pipelines, leading to incorrect trajectory predictions or behavioral estimations for robots and autonomous vehicles. Here, we evaluate whether the proposed ghost removal framework improves the detection accuracy of 3D object detectors.

\noindent\textbf{Experimental Settings.~}
We collected a test dataset for 3D object detection using the same sensor configuration as the Ghost-FWL dataset. The data were captured in outdoor and indoor environments containing glass surfaces, such as building entrances and glass-walled sidewalks, consisting of 102 frames and 239 object instances. We annotated bounding boxes for object instances, and labeled ghost points in the resulting point clouds.

\noindent\textbf{Metrics.~}
We evaluate ghost removal effectiveness by the Ghost False Positive (Ghost FP) Rate, the percentage of ghosts reflection incorrectly detected as pedestrians. Following~\cite{kraus2021radar}, we count a detection as a false positive when a bounding box overlaps with annotated ghost regions and is classified as a pedestrian.

\noindent\textbf{Comparative Methods.~}
We utilize the same LiDAR processing methods as \S\ref{subsec:slam-evaluation}. All methods employed the same 3D object detection model, PV-RCNN~\cite{Shi_PV-RCNN_Point-Voxel_Feature_Set_Abstraction_CVPR_2020}, pretrained on the KITTI dataset~\cite{geiger2013kitti}.

\begin{figure}[tp]
  \begin{center}
    \includegraphics[width=1.0\linewidth]{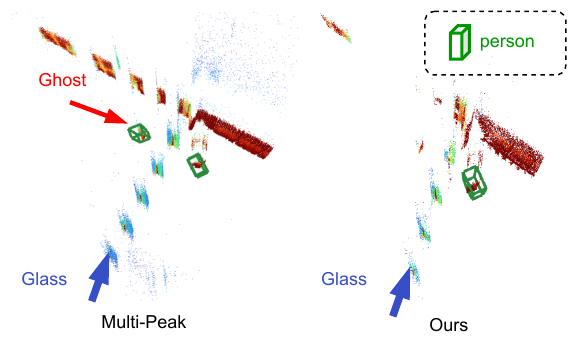}
    \vspace{-6mm}
    \caption{Qualitative evaluation of 3D object detection with Multi-Peak processing (left) and our ghost removal (right). Green bounding boxes indicate persons. With Multi-Peak, a ghost person is detected behind the glass wall, whereas our method suppresses this false detection.    
    }
    \label{fig:object_detection_result}
  \end{center}
  \vspace{-4mm}
\end{figure}

\noindent\textbf{Results.~}
Table~\ref{tab:detection_comparison} presents the quantitative results. Compared to the baseline LiDAR processing, our method achieves a 50$\times$ reduction in Ghost FP Rate (from 67.9\% to 1.34\%). 
~\cref{fig:object_detection_result} shows that ghosts of pedestrians appearing behind glass surfaces were effectively removed, enabling the detector to correctly identify only real objects.

\begin{table}[t]
\centering
\caption{Ghost-induced false positives in ghost detection.}
\label{tab:detection_comparison}
\setlength{\tabcolsep}{3pt} %
\renewcommand{\arraystretch}{0.95} %
\begin{tabularx}{0.48\textwidth}{l*{3}{>{\centering\arraybackslash}X}}
\toprule
Method & Ghost FP Rate [\%]~($\downarrow$) \\
\midrule
Dual-Peak~\cite{ouster,at128}  & 75.8 \\
Multi-Peak~\cite{avia} & 67.9  \\
\midrule
Ours & \textbf{1.34}\\
\bottomrule
\end{tabularx}
\vspace{-2mm}
\end{table}

\section{Conclusions and Future Work}  
\label{sec:conclusion}

This work introduces Ghost-FWL, the first large-scale annotated dataset for full-waveform ghost detection. Spanning 10 diverse scenes with over 24,000 annotated frames and 7.5 billion peak-level labels (100$\times$ larger than prior work), Ghost-FWL provides annotations for each reflection's physical origin: Ghost, Glass, or Object. By exploiting temporal intensity profiles invisible in standard point clouds, Ghost-FWL enables learning-based ghost removal that directly leverages multi-path reflection characteristics.
Models trained on Ghost-FWL significantly improve LiDAR-based SLAM and 3D object detection, contributing to enhanced safety in robotics and autonomous driving. 
We publicly release the dataset, code, and benchmarks to foster further research in robust LiDAR perception.

Our peak-level annotations focus on static multi-viewpoint captures to ensure high labeling quality; continuous mobile sequences are included for self-supervised pre-training but remain unlabeled. Extending annotations to these sequences would further benefit tracking and temporal modeling. Additionally, while our dataset emphasizes glass-induced ghosts in clear weather (the most common urban scenario), investigating other reflective materials (e.g. water, polished metals) and adverse conditions (e.g. rain, fog) would enable more comprehensive multi-path understanding.

\noindent\textbf{Acknowledgment.}
This research was supported in part by JST Next-generation Edge AI Semiconductor
JPMJES2515, JST CREST JPMJCR21D2, JPMJCR23M4, JST PRESTO JPMJPR22PA, JST ASPIRE JPMJAP2515, JSPS KAKENHI 24K02940, 	24K22296, and 25H01159.

\pagestyle{plain} %

\clearpage
\appendix
\section*{Supplementary Material}
\addcontentsline{toc}{section}{Supplementary Material}
{
\hypersetup{linkcolor=black}
\tableofcontents
}

\section{Overview of Supplementary Material}
This supplementary material provides additional details and further experimental results that complement the content presented in the main paper.
Please also refer to the \textbf{supplementary video} at~\url{https://keio-csg.github.io/Ghost-FWL/}, which presents the SLAM results of the comparative methods and our proposed method.
For clarity, we use \textcolor{red}{red} to denote references corresponding to the main paper, and \textcolor{cvprblue}{blue} to denote those corresponding to these supplementary materials.

\section{Ghost-FWL Dataset}
This section describes the details of the Ghost-FWL dataset.
\cref{fig:scene} shows an overview of all scenes in the Ghost-FWL dataset. Table \ref{tab:dataset-statistics} shows the number of frames for each scene.

\section{Fundamentals of LiDAR and Full-Waveform Signals}

LiDAR measures the three-dimensional structure of the environment by emitting laser pulses and computing the time-of-flight (ToF) of their returns. In standard commercial LiDAR systems, the raw received signal is internally processed to detect only the dominant return peaks, and the sensor outputs a set of 3D points, commonly referred to as a point cloud. This point-cloud representation is compact, easy to handle in downstream perception pipelines, and greatly reduces data bandwidth. However, this conversion discards most of the physical information originally present in the raw waveform, including material-dependent reflectance behavior, multi-path components, and the detailed temporal shape of each peak.

\begin{figure*}[t]
  \begin{center}
    \includegraphics[width=0.85\textwidth]{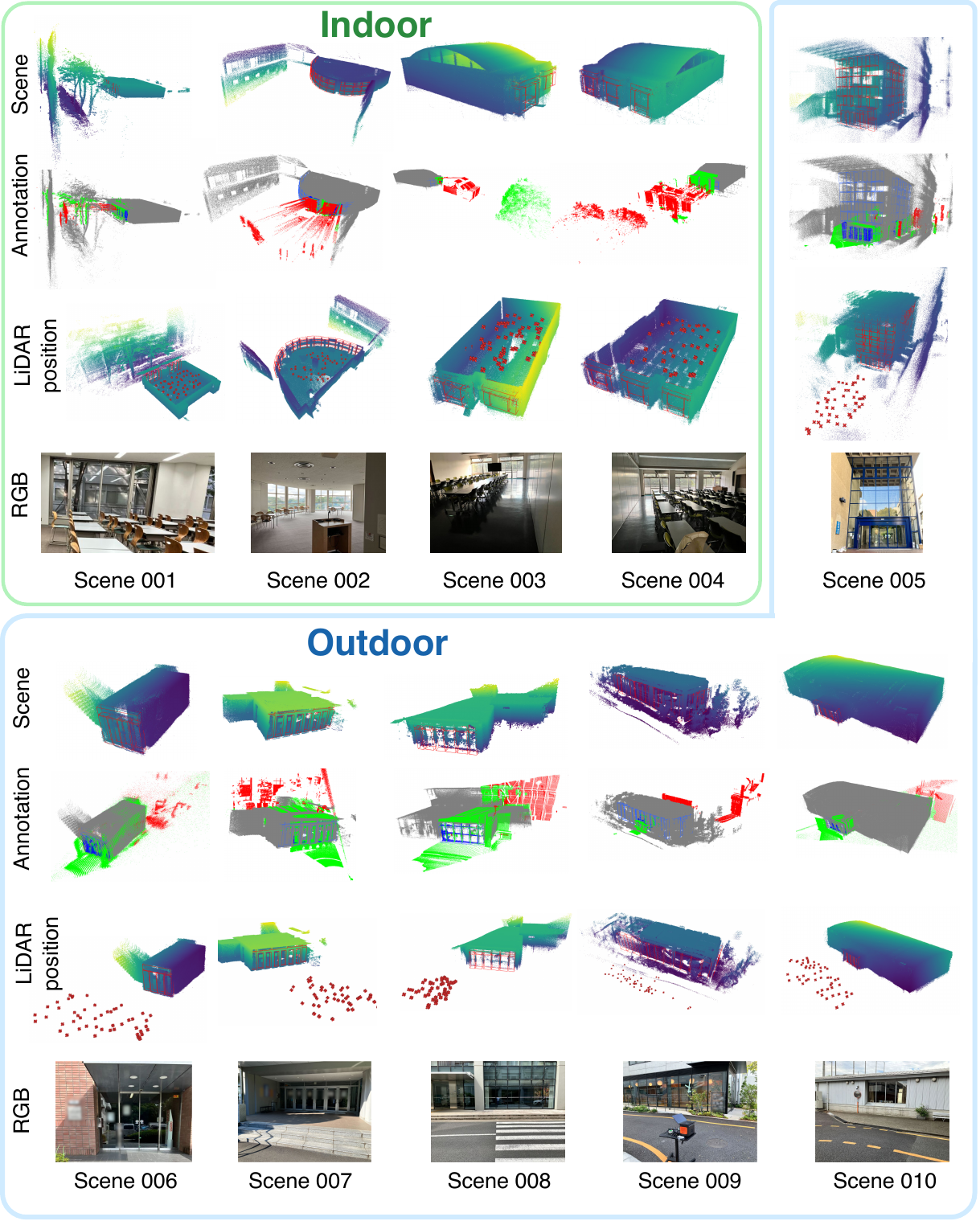}
    \caption{\textbf{All scenes in Ghost-FWL.} Our dataset includes both indoor and outdoor scenes. Based on the dense 3D maps as shown in Scene, we annotated FWL data with semantic labels: \textit{Ghost} (red), \textit{Object} (green), \textit{Glass} (blue), \textit{Noise}. Gray regions are excluded from annotation. Red crosses indicate the LiDAR positions during data acquisition. The RGB images show the scenery of the capture locations.}
    \label{fig:scene}
  \end{center}
\end{figure*}

In contrast, a Full-Waveform (FW) LiDAR records the complete temporal intensity profile of the returned signal for each beam direction. Because the full waveform preserves the full time evolution of the reflected light, it retains rich physical cues such as variations induced by material properties, surface geometry, incidence angle, and multi-path reflections through glass or other reflective structures including colored glass~(Scene~008 glass in \cref{fig:scene}) and film-covered glass~(Scene~002 glass in \cref{fig:scene}) surfaces.  
These temporal characteristics, suppressed or entirely lost in conventional point-cloud outputs, are crucial for analyzing and identifying ghost reflections, making FW LiDAR fundamentally more informative for ghost detection and removal.

\begin{table}[t]
  \centering
  \caption{{Ghost-FWL statistics by scene.}}
  \label{tab:dataset-statistics}
  \sisetup{group-separator=,}
  \renewcommand{\arraystretch}{1.15}
  \begin{tabularx}{\linewidth}{l*{2}{>{\centering\arraybackslash}X}}
    \toprule
    Scene & {Frames} & Location  \\
    \midrule
    001 & 2500  & Indoor  \\
    002 & 2500  & Indoor \\
    003 & 2749  & Indoor  \\
    004 & 1853  & Indoor  \\
    005 & 2500  & Outdoor   \\
    006 & 2445  & Outdoor  \\
    007 & 2461  & Outdoor \\
    008 & 2300  & Outdoor \\
    009 & 2354  & Outdoor   \\
    010 & 2750  & Outdoor \\
    \midrule
    \textbf{TOTAL} & \textbf{24412} &  Indoor 4 / Outdoor 6  \\
    \bottomrule
  \end{tabularx}
\end{table}

\subsection{Annotation Strategy}

Annotating FWL data that contain ghost reflections are fundamentally challenging. Ghost returns are virtual reflections that may not correspond to any physical surface in the scene; therefore, their spatial location, intensity, and temporal patterns vary depending on the geometry and reflectance of glass or other reflective materials. As a result, no direct ground-truth reference exists for identifying where ghost peaks should appear. In addition, raw FWL data include a large amount of noise originating from natural illumination such as sunlight, as well as weak background reflections. These noise peaks occur randomly along the temporal axis and cannot be separated from ghost peaks by simple filtering or thresholding, making conventional annotation approaches unreliable.

To address these difficulties, we construct an annotation framework that jointly leverages a high-precision 3D map (GT) and accumulated FWL data.
For each scene, we prepare
(i) a high-accuracy 3D map, and
(ii) multiple FWL frames obtained from the same environment.
The FWL frames are first accumulated to obtain a high-SNR waveform, which suppresses stochastic noise while enhancing stable reflection components, including both object returns and consistent multi-path signals associated with ghost reflections. The accumulated waveform is then converted into a point cloud and compared with the GT map to determine whether each FWL peak corresponds to a real object surface, a glass region, a ghost reflection, or noise. Because ghost reflections typically appear at locations that deviate from the real-world geometry, the spatial discrepancy between the accumulated FWL-derived point cloud and the GT map provides a reliable and discriminative cue for identifying ghost peaks.

Labels are first assigned in the point-cloud domain and then transferred back to the corresponding peaks in the accumulated FWL data, yielding peak-level annotations suitable for supervised learning.

\subsection{Annotation Pipeline}
\begin{figure}[t]
  \begin{center}
    \includegraphics[width=1.0\linewidth]{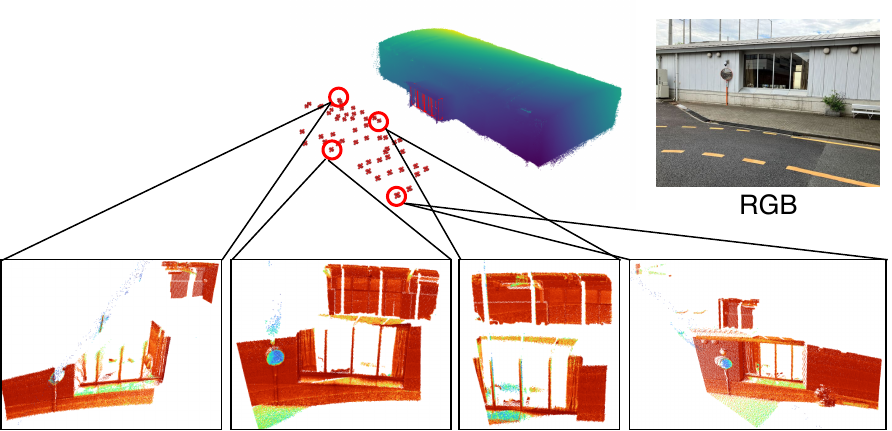}
    \caption{Example of LiDAR position and perspective.}
    \label{fig:capture_view}
  \end{center}
  \vspace{-4mm}
\end{figure}

\begin{enumerate}
\item \textbf{GT Map Construction:}
A high-precision 3D map is constructed using a commercial LiDAR sensor (Livox Mid-360~\cite{mid-360}) together with the SLAM algorithm fastlio2~\cite{xu2022fast}.
After generating the map, ghost points and noise points are removed, and two spatial regions are manually defined using labelCloud~\cite{Sager_2022}:
the \textit{glass region} $\mathcal{G}$, which indicates where glass surfaces are likely to exist, and the \textit{reflection region}~$\mathcal{R}$, which denotes a larger zone extending radially behind the reflective surfaces where multi-path reflections may occur.
Because the alignment between the FWL-derived points and the GT map is performed manually, small alignment errors are unavoidable. To prevent these errors from negatively affecting the labeling process, $\mathcal{R}$ is intentionally defined to be larger than $\mathcal{G}$.

\item \textbf{Alignment with Accumulated FWL Data:}
For each scene, 37--55 viewpoints are selected, and approximately 50 FWL frames are captured per viewpoint.
\cref{fig:capture_view} shows example of LiDAR positions and viewpoints.
These frames are accumulated to obtain a high-SNR FWL signal.
Peak detection is then applied to the accumulated waveform, and the detected peaks are converted into a point cloud.
The accumulated FWL point cloud is manually aligned with the GT map to correct for discrepancies between their coordinate systems.

\item \textbf{Label Assignment:}
Let $\mathcal{M}$ denote the set of GT map points.
For each accumulated FWL point $\mathbf{x}$, we compute its nearest-neighbor distance to the GT map as
\begin{equation}
        d(\mathbf{x}) = \min_{\mathbf{y}\in\mathcal{M}} \|\mathbf{x}-\mathbf{y}\|
\end{equation}

Using this distance and the region definitions, each point is classified as follows:

\begin{itemize}
        \item \textbf{Glass}: 
        \begin{equation}
            \mathbf{x} \in \mathcal{G}
        \end{equation}
        \item \textbf{Object}:
        \begin{equation}
            d(\mathbf{x}) < \tau
            \quad \text{and} \quad 
            \mathbf{x} \notin \mathcal{G}
        \end{equation}
        \item \textbf{Ghost}:
        \begin{equation}
            d(\mathbf{x}) > \tau
            \quad \text{and} \quad
            \mathbf{x} \in \mathcal{R}
        \end{equation}
        \item \textbf{Noise:}
        All remaining points that do not satisfy any of the above conditions.
\end{itemize}

Here, $\tau$ denotes the Euclidean distance threshold.
Once these labels are assigned in the point-cloud domain, they are transferred back to the corresponding peaks in the accumulated FWL data, completing the peak-level annotation procedure.
The parameters used to construct the Ghost-FWL dataset are summarized in Tab.~\ref{tab:annotation-param}.
\end{enumerate}

\begin{table}[t]
\centering
\caption{Annotation parameters for each scene in the Ghost-FWL dataset. 
The threshold $\tau$ is the nearest-neighbor distance used to distinguish Object and Ghost points, and ``\# Glass Areas'' denotes the number of manually annotated glass regions $\mathcal{G}$.}
\label{tab:annotation-param}
\setlength{\tabcolsep}{4pt}
\renewcommand{\arraystretch}{0.95}
\begin{tabularx}{0.48\textwidth}{l*{2}{>{\centering\arraybackslash}X}}
\toprule
Scene & Threshold $\tau$ & \# Glass Areas \\
\midrule
001 & 0.5 & 8 \\
002 & 0.5 & 48 \\
003 & 0.5 & 10 \\
004 & 0.5 & 9 \\
005 & 0.5 & 15 \\
006 & 0.5 & 5 \\
007 & 0.5 & 17 \\
008 & 0.5 & 12 \\
009 & 0.5 & 15 \\
010 & 0.5 & 4 \\
\bottomrule
\end{tabularx}
\vspace{-4mm}
\end{table}

\label{sec:lidar-stack}

\subsection{Improving Annotation Quality via Waveform-Level Accumulation}

The raw full-waveform signal recorded by a LiDAR sensor inevitably contains a considerable amount of noise. Such noise appears as randomly distributed peaks along the temporal axis and is primarily caused by external illumination, including sunlight, as well as weak background reflections from the ground and surrounding surfaces. To suppress these stochastic components, modern LiDAR systems employ an accumulation mechanism in which multiple laser pulses are emitted in the same direction and their corresponding waveforms are aggregated. Since true reflections from physical objects consistently appear at the same temporal position across pulses, while noise peaks vary randomly, accumulation enhances stable reflection components and averages out random noise.

\begin{figure}[t]
  \begin{center}
    \includegraphics[width=0.9\linewidth]{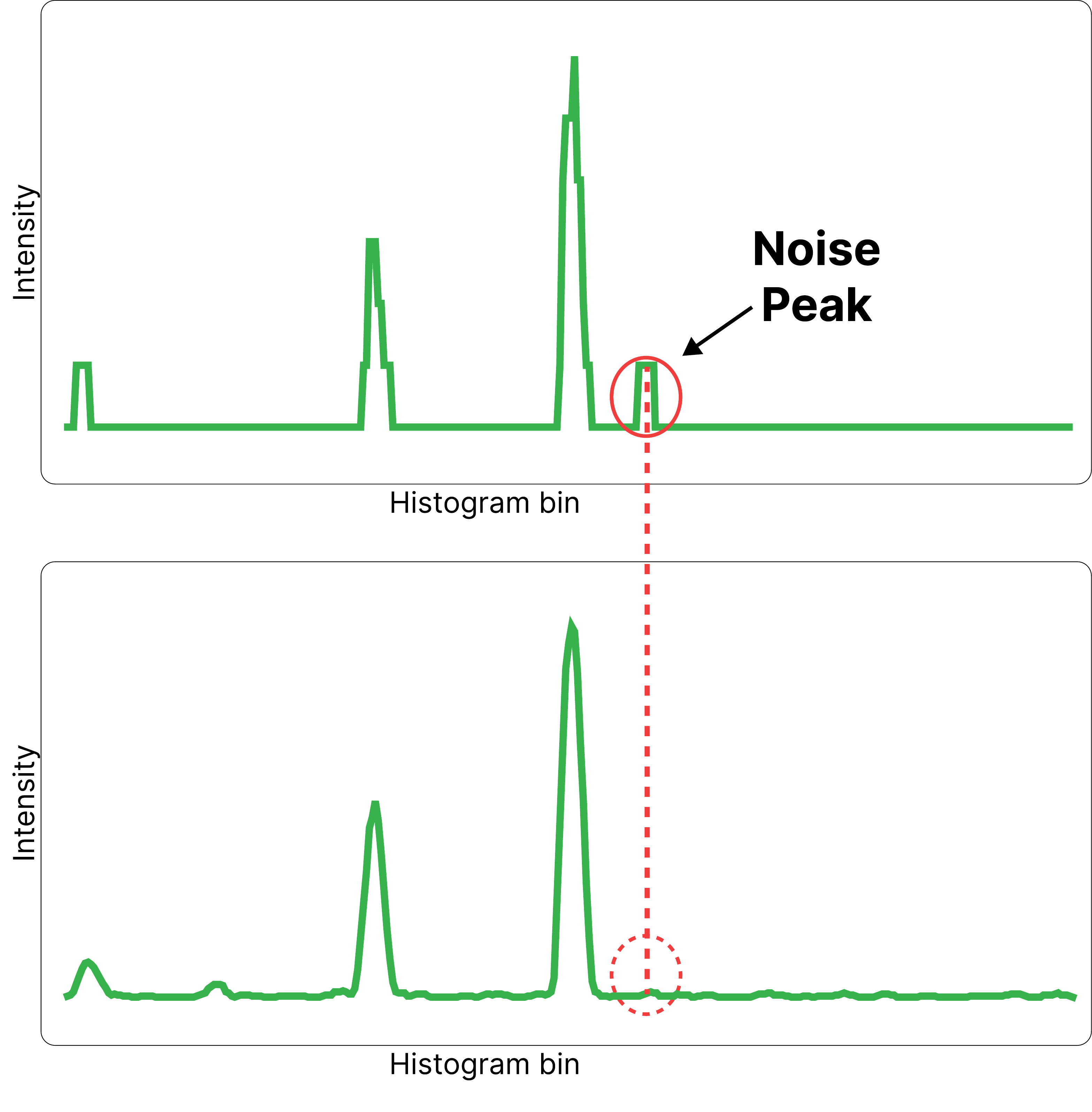}
    \vspace{-4mm}
    \caption{{Example of 1 frame FWL data(top) and 50$\times$ accumulation FWL data  (bottom).}
    }
    \label{fig:annotation_histogram}
  \end{center}
  \vspace{-4mm}
\end{figure}

We leverage the same accumulation strategy to improve the reliability of our annotation pipeline. As shown in ~\cref{fig:annotation_histogram}, for each viewpoint, 50 FWL frames captured from the same location are aggregated to generate a high-SNR waveform. This process amplifies not only direct reflections from real objects but also consistent multi-path components responsible for ghost reflections, while effectively suppressing sporadic noise induced by sunlight or ground scattering. The accumulated waveform is then converted into a point cloud, ensuring that only physically meaningful reflection peaks remain. This high-quality representation provides a robust basis for distinguishing true objects from ghost reflections, enabling more reliable detection and highly accurate peak-level annotation of ghost returns.

\begin{figure}[t]
  \begin{center}
    \includegraphics[width=0.9\linewidth]{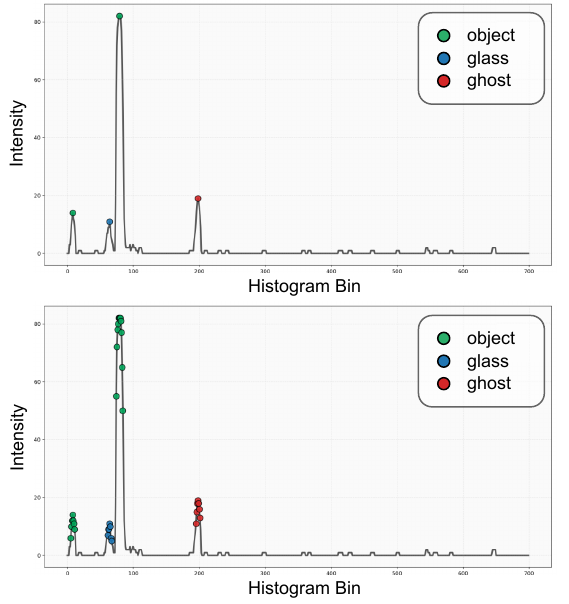}
    \vspace{-2mm}
    \caption{{Example of raw annotation (top) and processed annotation for training (bottom).
    The circles indicate the annotation points for each class: \textit{Object} (green), \textit{Glass} (blue) and \textit{Ghost} (red).}}
    \label{fig:annotation_histogram_training}
  \end{center}
\end{figure}

\begin{table*}[t]
\centering
\caption{FWL-MAE Architecture.}
\label{tab:architecture}
\renewcommand{\arraystretch}{1.2}
\begin{tabular}{l|c|c}
\hline
\textbf{Name} & \textbf{Layer setting} & \textbf{Output dimension} \\
\hline\hline

\multirow{2}{*}{Patch Embedding}
 & 3D Conv 
 & \multirow{2}{*}{$B \times N_{\text{patch}} \times D_{\text{encoder}}$} \\
 & Flatten + Transpose 
 &  \\

\hline

{Positional Encoding}
 & Sinusoidal Position Encoding (PE)
 & {$B \times N_{\text{patch}} \times D_{\text{encoder}}$} \\

\hline

{Masking}
 & Select Unmasked Tokens + PE
 & {$B \times N_{\text{unmasked}} \times D_{\text{encoder}}$} \\

\hline\hline

\multirow{2}{*}{Encoder}
 & Transformer Blocks $\times$ 6
 & (pretraining) $B \times N_{\text{unmasked}} \times D_{\text{encoder}}$  \\
 & LayerNorm 
 & (fine-tuning) $B \times N_{\text{patch}} \times D_{\text{encoder}}$   \\

\hline\hline

\multirow{4}{*}{Concatenation Tokens}
 & Linear Projection [$D_{\text{encoder}} \rightarrow D_{\text{decoder}}$] & \multirow{4}{*}{$B \times N_{\text{patch}} \times D_{\text{decoder}}$} \\
 & Unmasked Features + PE  & \\
 & Masked Tokens + PE & \\
 & Concat (Unmasked $N_{\text{unmasked}}$, Masked $N_{\text{masked}}$) & \\

\hline\hline

\multirow{2}{*}{Peak Position Head}
 & Linear [$D_{\text{decoder}} \rightarrow K$] 
 & \multirow{2}{*}{$B \times N_{\text{patch}} \times K$} \\
 & Sigmoid $\times$ ($T$ - 1) & \\

\hline

\multirow{2}{*}{Peak Width Head}
 & Linear [$D_{\text{decoder}} \rightarrow K$] 
 & \multirow{2}{*}{$B \times N_{\text{patch}} \times K$} \\
 & Softplus & \\

\hline

\multirow{2}{*}{Peak Height Head}
 & Linear [$D_{\text{decoder}} \rightarrow K$] 
 & \multirow{2}{*}{$B \times N_{\text{patch}} \times K$} \\
 & Softplus & \\

\hline\hline

\multirow{2}{*}{Decoder}
 & Transformer Blocks $\times$ 6
 & \multirow{2}{*}{$B \times N_{\text{mask}} \times (H_{\text{patch}} \times W_{\text{patch}} \times T_{\text{patch}})$} \\
 & Linear [$D_{\text{decoder}} \rightarrow (H_{\text{patch}} \times W_{\text{patch}} \times T_{\text{patch}})$] & \\

\hline\hline

\multirow{3}{*}{Classification Head}
 & Linear [$D_{\text{encoder}} \rightarrow \frac{D_{\text{encoder}}}{2}$] + ReLU + Dropout
 & \multirow{3}{*}{{$B \times C \times T \times H \times W$}} \\
 & Linear [$\frac{D_{\text{encoder}}}{2} \rightarrow (H_{\text{patch}} \times W_{\text{patch}} \times T_{\text{patch}} \times C)$]
 & \\
 & Reshape Patches $\rightarrow$ FWL data \\

\hline

\end{tabular}

\end{table*}

\subsection{Annotation Processing for Training}
During training, both the input FWL data and the annotations must be down-sampled due to GPU memory limitations.
However, peak annotations exist only at the single point at the exact peak of the histogram, so they can be lost during down-sampling along the $T$-axis when the sampling interval is large.
Therefore, during training, we expand each annotation around the original peak position by its full width at half maximum (FWHM). 
This expansion ensures that, the ground-truth peak annotation still covers the peak of the same waveform even after down-sampling.
The comparison between raw and processed annotations for training is shown in \cref{fig:annotation_histogram_training}.

\section{Implementation Details of the FWL-based Ghost Removal Framework}
This section presents the implementation details of the FWL-based Ghost Removal Framework.
Table \ref{tab:architecture} shows a detailed architecture of FWL-MAE.
All training and processing were performed on a computer equipped with an Intel Xeon w5-3535X CPU and a single RTX 6000 Ada Generation GPU, running the Ubuntu 22.04 operating system.

\subsection{Full Waveform LiDAR Masked Autoencoder}
\label{subsec:fwl-mae}
This subsection describes the implementation details of the self-supervised pretraining method, the Full-Waveform LiDAR Masked Autoencoder (FWL-MAE), which is designed to obtain latent representations of histograms from FWL data.

\noindent\textbf{Model.~}
The implementation of the Transformer-based baseline model is inspired by VideoMAE~\cite{tong2022videomae}.
We first apply 3D convolutions to the FWL data to generate patch embeddings $(B, N_{\text{patch}}, D_{\text{encoder}})$, where $B$ is the batch size, $N_{\text{patch}}$ is the number of patches, and $D_{\text{encoder}}$ is the encoder embedding dimension. 
For masking, we randomly sample spatial patches in the $(x, y)$ region and mask all temporal bins along the $T$ axis within each selected patch.

Only the unmasked patches are fed into a Transformer encoder composed of six Transformer blocks with six attention heads in each block.
The Transformer Encoder outputs feature vectors for the unmasked regions with shape $(B, N_{\text{unmasked}}, D_{\text{encoder}})$, where $B$ is the batch size, $N_{\text{unmasked}}$ is the number of unmasked patches, and $D_{\text{encoder}}$ is the encoder embedding dimension.

The Transformer decoder consists of six Transformer blocks with six attention heads in each block.
It takes as input the feature vectors of the unmasked regions and the mask tokens $(B, N_{\text{patch}}, D_{\text{decoder}})$. The decoder then reconstructs the FWL data corresponding to the masked patches.

In the Peak Head, the peak \textit{position} ($p$), \textit{amplitude} ($a$), and \textit{width} ($w$)  of the FWL data are estimated simultaneously.
The ground-truth peak position, amplitude, and full width at half maximum (FWHM) are extracted directly from the input FWL data.
Under the assumption that peaks farther from the LiDAR sensor are less reliable and less informative, the Peak Head estimates only the $K$ peaks, where $K$ is set to 4 in this paper.
If fewer than $K$ peaks exist in input FWL data, the missing ground-truth peak position, amplitude, and width are padded with zeros. Since peak prediction is performed at the patch level, the ground-truth peak parameters for each patch are obtained by averaging the peak values within the corresponding spatial patch region.

In our setting, the masking ratio of FWL data was set to $70\%$. The input FWL data size is $(H, W, T) = (128, 128, 256)$ and the patch size is $(H_{\text{patch}}, W_{\text{patch}}, T_{\text{patch}}) = (16, 16, 256)$. The embedding dimension is $D_{\text{encoder}} = 768, D_{\text{decoder}}=384$.

\noindent\textbf{Preprocessing.~}
The raw FWL data were reshaped to $(H, W, T) = (128, 128, 256)$ before being fed into the model through the following preprocessing. First, the top and bottom 90 bins corresponding to reflections from the ceiling and floor and the front 25 bins containing noise from internal sensor reflections were removed. 
The remaining data were then uniformly down-sampled along the $T$ axis and randomly cropped into $H \times W$ size.
The preprocessed FWL data is then fed into the model. 

\noindent\textbf{Training.~}
We performed pretraining with FWL-MAE using 8,933 unlabeled frames captured in a mobile environment.
We used AdamW~\cite{Loshchilov2017DecoupledWD} as optimizer.The AdamW hyperparameters were set to $\beta_1=0.9, \beta_2=0.999$, $\epsilon=1\times10^{-8}$, weight decay of $\lambda=1 \times 10^{-2}$, and a learning rate of $\alpha = 1 \times 10^{-3}$.
The batch size was set to 32, and training was performed for 100 epochs.
The weighting coefficients for the loss function $\mathcal{L}_{\text{FWL-MAE}}$ are $\lambda_p= 1.0$, $\lambda_a= 1.0$, and $\lambda_w = 0.5$.

\subsection{Ghost Detection and Removal}
\label{subsec:ghost-detection-and-removal}
This subsection describes the method for peak-level classification for ghost detection and removal.
To detect and remove ghosts, our method takes the FWL data $\bm{V} \in \mathbb{R}^{H \times W \times T}$ as input and estimates the class probabilities $\bm{P} \in \mathbb{R}^{H \times W \times T \times C}$ for the categories \textit{Glass}, \textit{Ghost}, \textit{Object}, and \textit{Noise}.

\noindent\textbf{Model.~}
To extract informative features from the FWL data, we use the encoder pretrained with FWL-MAE and keep its weights frozen to obtain latent representations $f_{\theta}(\bm{V})$.
A lightweight classification head composed of two linear layers is then applied to predict the class probabilities for all FWL data coordinates.

\noindent\textbf{Preprocessing.~}
The input size to the model and the preprocessing procedure are the same as those used during the pretraining with FWL-MAE.

\noindent\textbf{Training.}
We used a data split of 13,853 for training, 2,994 for validation, and 1,427 for testing. 
The training and validation sets contain data captured in Scene 001, 003, 004, 005, 006, 008 and 010.
The testing set contains data captured in Scene 002, 007 and 009.
Although the training and validation sets were captured in the same scenes, no overlapping frames were used.
The test set consists solely of unseen scenes that are not included in either the training or validation data.
In this study, we used the above split, but it can be modified as needed.

We used AdamW~\cite{Loshchilov2017DecoupledWD} as optimizer.The AdamW hyperparameters were set to $\beta_1=0.9, \beta_2=0.999$, $\epsilon=1\times10^{-8}$, weight decay of $\lambda=1 \times 10^{-2}$, and a learning rate of $\alpha = 1 \times 10^{-3}$.
The batch size was set to 32, and training was performed for 100 epochs.

\noindent\textbf{Loss function.~}We adopt the focal loss~\cite{lin2017focal}, which mitigates the impact of class imbalance in multi-class classification.
Our task poses a challenging classification problem
involving highly imbalanced data, where minority classes such as \textit{Ghost} coexist with the majority \textit{Noise} class.
The loss function is defined as follows:
\begin{equation}
\mathcal{L_{\text{Focal}}}= -\sum_{c=1}^{C}
\alpha_{c} (1 - \bm{p}_{c})^{\gamma} \log \bm{p}_{c},
\end{equation}
where $C$ is the number of class, $\bm{p}_{c}$ denotes the predicted probability for the ground-truth class $c$.
$\alpha_c$ is a weighting factor for each class (Glass, Ghost, Object, Noise), used to compensate for class imbalance,
and $\gamma$ is the focusing parameter that controls the trade-off between easy and hard samples.
The number of classes are set as  $C = 4$ (Glass, Ghost, Object, and Noise). The parameters were set to $\alpha_{\text{glass}} = 0.25$, $\alpha_{\text{ghost}} = 0.7$, $\alpha_{\text{object}} = 0.05$, $\alpha_{\text{noise}} = 0.0001$, and $\gamma = 2.0$.

\noindent\textbf{Inference.~}
We describe the inference procedures for downstream tasks such as SLAM and object detection, as well as for qualitative evaluation of classification.

First, as in the training phase, the top and bottom 90 bins corresponding to reflections from the ceiling and floor and the front 25 bins containing noise from internal sensor reflections were removed.

During inference, we use FWL data with the same size $(H, W, T)=(128,128,256)$ as those used during training.
The raw data are down-sampled along the $T$-axis.
Next, unlike in training where random cropping is applied, the FWL data are cropped sequentially starting from the patch coordinate $(x, y) = (0,0)$
The FWL data are then processed sequentially using a sliding window applied to cropped regions that do not overlap.
If the window extends beyond the valid range, the outside area is padded with zero.
The inference results obtained sequentially through the sliding window process are then merged, and finally up-sampled to match the original input shape.
During up-sampling, zeros are padded between values to prevent the number of predicted classes from artificially increasing.
During inference, the predicted class was determined as the one with the highest probability if it exceeded 0.5; otherwise, it was assigned to \textit{Undefined}.

\section{Experiments and Results}
This section summarizes additional experiments that could not be included in the main paper.

\subsection{Ghost Denoising Evaluation}

\noindent\textbf{Detail of Metrics.~} We evaluate ghost detection at two levels: peak-level and point-level. For peak-level evaluation, we follow Scheuble et al.~\cite{scheublelidar} and report recall, measuring the proportion of correctly detected ghost peaks among all ground-truth ghost peaks. 
In Recall, only the peaks within the FWL data are detected, and evaluation is performed on their positions.

For point-level evaluation, we introduce the \textit{Ghost Removal Rate}, which measures the proportion of ghost points successfully removed after converting predicted peaks to 3D point clouds. This metric is inspired by the snow removal rate in Charron et al.~\cite{charron2018noising} and assesses practical denoising effectiveness in downstream tasks.
For each GT point, if no point in the denoised points was found within a radius $r$, the GT point was counted as removed. The Removal Rate is then defined as the ratio of removed GT points to the total number of GT points:
\begin{equation}
\text{Ghost Removal Rate} = \frac{N_{\text{removed}}}{N_{\text{GT}}},
\end{equation}
where $N_{\text{removed}}$ denotes the number of removed ghost points, and $N_{\text{GT}}$ is the total number of GT points. The radius was set to $r=0.001$ in meters.

\subsubsection{Ghost Classification Evaluation}
\label{sssec:classification}

\noindent\textbf{Metrics.~}
We follow Scheuble et al.~\cite{scheublelidar} and report recall, measuring the proportion of correctly detected ghost peaks among all ground-truth ghost peaks. 

\noindent\textbf{Comparative Methods.~}
To investigate the effectiveness of the proposed FWL-MAE pretrained encoder, we compared five models: (1) the proposed model incorporating FWL-MAE (\textbf{Ours}), (2) the model without FWL-MAE (\textbf{Ours w/o FWL-MAE}), (3) the model pretrained using a general MAE designed for transient imaging (\textbf{MARMOT}~\cite{shen2025marmot}), (4) the model for transient imaging (\textbf{Lindell et al.}~\cite{lindell2018single}) and (5) the most commonly used 3D convolution-based model (\textbf{3D U-Net~\cite{cicek2016unet3d}}.

For (1), the proposed method applied self-supervised pretraining using FWL-MAE and then finetuned the model on the Ghost-FWL dataset.
For (2), the model was trained from scratch on the Ghost-FWL dataset without FWL-MAE, using random weight initialization.
For (3), the same architecture was pretrained with MARMOT~\cite{shen2025marmot}
, a general MAE design for transient imaging, and then finetuned on the Ghost-FWL dataset.
For (4), Lindell et al.~\cite{lindell2018single} is a 3D convolution-based depth estimation model takes transient imaging as input. Although it can also utilize intensity image, we use only the transient input in our experiments.
For (5), 3D U-Net~\cite{cicek2016unet3d} is employed as a commonly used 3D convolution-based method capable of handling 3D inputs. 

\noindent\textbf{Results.~}
As shown in Table \ref{tab:ghost_class_evaluation_supp}, the proposed Transformer-based model with FWL-MAE achieves superior performance in ghost-detection recall compared with models using alternative pretraining strategies as well as existing 3D convolution-based approaches. These results demonstrate the effectiveness of incorporating FWL-MAE, which enables pretraining that more effectively captures the physical characteristics of FWL data.

\begin{figure*}[t]
  \begin{center}
    \includegraphics[width=1.0\textwidth]{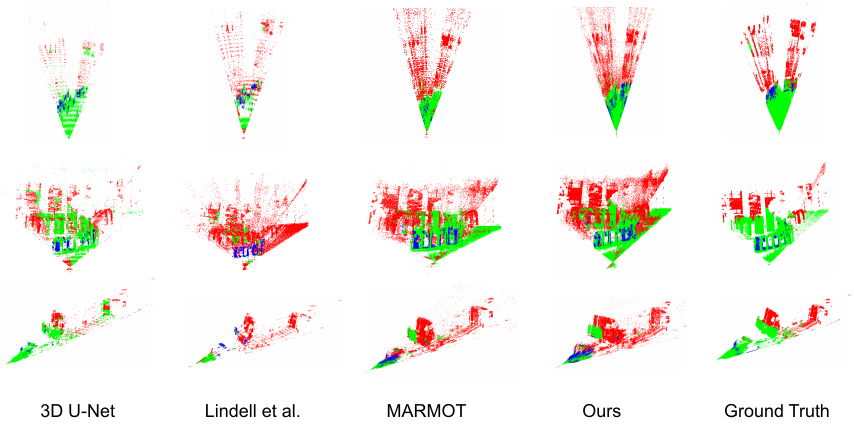}
    \caption{Peak classification results.
    Red, green and blue indicates \textit{Ghost}, \textit{Object} and \textit{Glass}, respectively.}
    \label{fig:classification}
  \end{center}
\end{figure*}

\cref{fig:classification} shows the classification results of peak in the FWL data. 
The prediction results of 3D convolution-based models, 3D U-Net~\cite{cicek2016unet3d} and Lindell et al.~\cite{lindell2018single}, contain many regions where the class labels are incorrectly estimated.
In contrast, the proposed Transformer-based models, pretrained with MARMOT~\cite{shen2025marmot} or FWL-MAE, achieve more accurate classification.
However, compared to the ground truth, they still produce a larger number of predictions labeled as ghost. This occurs because noise peak positions are sometimes classified as ghost. Although misclassifying noise as ghost has limited impact on downstream tasks, further improvements in classification accuracy remain desirable.

\begin{table}[t]
\centering
\caption{Comparison of ghost removal performance with other methods.}
\vspace{-2mm}
\label{tab:ghost_class_evaluation_supp}
\setlength{\tabcolsep}{6pt}
\renewcommand{\arraystretch}{1.0}

\begin{tabularx}{0.9\linewidth}{l>{\centering\arraybackslash}X}
\toprule
Method & Recall ($\uparrow$) \\
\midrule
3D U-Net~\cite{cicek2016unet3d} & 0.391 \\
Lindell et al.~\cite{lindell2018single} & 0.641 \\
MARMOT~\cite{shen2025marmot} & 0.746 \\
\midrule
Ours w/o FWL-MAE  & 0.704 \\
Ours & \textbf{0.751} \\
\bottomrule
\end{tabularx}
\end{table}

\subsubsection{Sensitivity analysis for classification threshold}

We conducted a sensitivity analysis on the ghost detection threshold, resulting in recall values of 0.751, 0.702, and 0.622 at thresholds of 0.5, 0.6, and 0.7, respectively.
The threshold of 0.5 was adopted for our main paper as it yielded the best performance.

\begin{table}[t]
\centering
\caption{{Ablation study of FWL-MAE.}}
\label{tab:ablation_fwl_mae}
\begin{tabularx}{0.9\linewidth}{
>{\centering\arraybackslash}p{1.2cm}ll
}
\toprule
Train Data & Method & Recall ($\uparrow$) \\
\midrule
\multirow{2}{*}{100\%} 
 & Ours w/o FWL-MAE & 0.704 \\
 & Ours             & 0.751 (+ 0.047) \\
\midrule
\multirow{2}{*}{70\%} 
 & Ours w/o FWL-MAE & 0.602 \\
 & Ours             & 0.692 (+ 0.090) \\
\midrule
\multirow{2}{*}{50\%} 
 & Ours w/o FWL-MAE & 0.403 \\
 & Ours             & 0.603 (+ 0.200) \\
\bottomrule
\end{tabularx}
\end{table}

\subsubsection{Ablation Study of FWL-MAE}
We conducted quantitative experiments on classification with varying amounts of training data to demonstrate the effectiveness of self-supervised pretraining with FWL-MAE.

\noindent\textbf{Experimental Settings.~}
When the amount of training data described in \S\ref{subsec:ghost-detection-and-removal} is treated as 100\%, we conducted additional experiments by reducing the training data to about 70\% and 50\%.
For the 70\% set contains data captured in Scene 001, 003, 004, 005 and 006, and for the 50\% set contains data captured in Scene 001, 004, 005, and 006.
The amount of test data was kept unchanged.

\noindent\textbf{Results.~}
Table \ref{tab:ablation_fwl_mae} shows the recall scores of the proposed method with FWL-MAE pretraining compared to the method without FWL-MAE, evaluated under different amounts of training data.
As the amount of training data decreases, the recall of the method without FWL-MAE drops substantially, while the proposed method maintains higher performance. Consequently, the performance gap between the two methods widens, highlighting  the strong effectiveness of FWL-MAE pretraining, particularly in fewer data settings.

\begin{table}[t]
\centering
\caption{Computational cost and inference time.
FPS$^\dagger$ denotes the inference speed for model input size $(H, W, T)=(128,128,256)$. 
FPS$^\ddagger$ denotes the inference for full-frame size $(H, W, T)=(332, 400, 256)$.}
\vspace{-2mm}
\label{tab:computational_cost}
{\small
\begin{tabularx}{\linewidth}{
l
>{\centering\arraybackslash}X
>{\centering\arraybackslash}X
>{\centering\arraybackslash}X
>{\centering\arraybackslash}X}
\hline
Method & Params [M] & FLOPs [G] & FPS$^\dagger$ & FPS$^\ddagger$ \\
\hline
3D U-Net~\cite{cicek2016unet3d} & 90.3 & 7610 & 6.6 & 0.45 \\
Lindell et al.~\cite{lindell2018single} & 1.8 & 4090 & 7.0 & 0.48 \\
Ours & 194.1 & 23.5 & \textbf{32.1} & \textbf{2.35} \\
\hline
\end{tabularx}
}
\vspace{-3mm}
\end{table}

\subsubsection{Computational cost and inference speed}
Table~\ref{tab:computational_cost} summarizes the computational costs of our method compared to existing 3D CNN-based methods evaluated in ~\cref{sssec:classification}. 
Our Transformer-based approach achieves higher computational efficiency than 3D CNN-based methods. 
This efficiency stems from our strategy of dividing the input into spatial patches while treating the temporal dimension as a single tube, thereby reducing the total number of operations. 
In contrast, 3D CNNs involve computationally expensive convolutions across both spatial and temporal dimensions, resulting in more operations and slower inference.
Although our method significantly improves throughput for the model input size (FPS$^\dagger$), the inference speed per frame (FPS$^\ddagger$) remains limited. 
This is primarily because our current framework requires iterative inference for each frame due to memory constraints and the high computational overhead of processing an entire frame at once.
Therefore, real-time performance remains challenging.

\subsubsection{Efficacy for reflective materials}
\begin{figure}[t]
  \begin{center}
    \includegraphics[width=1.0\linewidth]{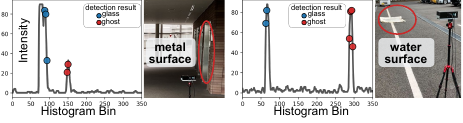}
    \vspace{-8mm}
    \caption{Our model's ghost detection on non-glass surfaces}
    \label{fig:histogram}
  \end{center}
\end{figure}

We conduct additional qualitative experiments on reflective materials other than glass, including water and metal surfaces, as shown in Fig.~\ref{fig:histogram}.
Although our model is primarily trained on glass-induced ghosts, it also successfully detects ghosts arising from water and metal surfaces, suggesting that it captures material-invariant FWL characteristics of multi-path reflections.
We further extend the Ghost-FWL dataset to include these additional surfaces.

\begin{table*}[t]
\centering
\caption{SLAM performance ablation with different point-cloud processing methods.}
\label{tab:slam_comparison_transposed}
\setlength{\tabcolsep}{3pt}
\renewcommand{\arraystretch}{1.05}
\begin{tabularx}{\textwidth}{l*{3}{>{\centering\arraybackslash}X}}
\toprule
 & Raw & Statistical Outlier Filter & Radius-based Outlier Filter \\
\midrule
Dual-Peak ATE [m] ($\downarrow$) 
    & 1.248$\pm$0.947
    & 0.639$\pm$0.573
    & 0.503$\pm$0.593 \\
Multi-Peak ATE [m] ($\downarrow$) 
    & 2.489$\pm$1.432 
    & 1.232$\pm$0.928
    & 0.887$\pm$1.079 \\
Ours ATE [m] ($\downarrow$) 
    & \textbf{0.294$\pm$0.244}
    & \textbf{0.248$\pm$0.171}
    & \textbf{0.328$\pm$0.303} \\
\midrule
Dual-Peak RTE [m] ($\downarrow$) 
    & 1.221$\pm$0.928
    & 0.608$\pm$0.529
    & 0.471$\pm$0.555 \\
Multi-Peak RTE [m] ($\downarrow$) 
    & 2.513$\pm$1.440 
    & 1.280$\pm$0.901
    & 0.832$\pm$1.015 \\
Ours RTE [m] ($\downarrow$) 
    & \textbf{0.288$\pm$0.237}
    & \textbf{0.243$\pm$0.157}
    & \textbf{0.319$\pm$0.291} \\
\bottomrule
\end{tabularx}
\end{table*}

\begin{figure}[t]
  \begin{center}
    \includegraphics[width=1.0\linewidth]{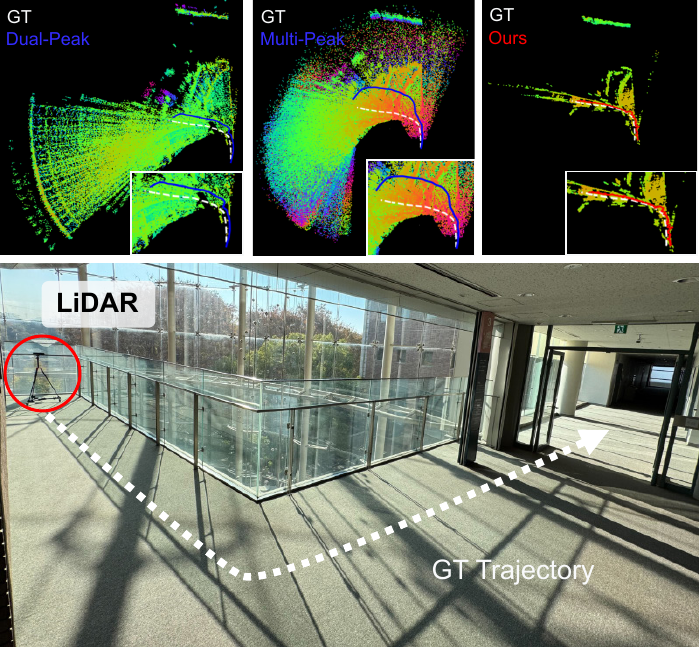}
    \caption{Trajectory and mapping results using Dual-Peak processing (left), Multi-Peak processing (center), and our ghost-removal method (right).
    The bottom image shows the scenery of the SLAM evaluation, recorded in a corridor enclosed by glass railings and doors.
     The scene is same to that used in the main paper Fig \textcolor{red}{5}, with the RGB scene image and the visualization results for the Dual-Peak method additionally included.
    }
    \label{fig:slam}
    \vspace{-4mm}
  \end{center}
\end{figure}

\subsubsection{Why Prior Ghost Removal Methods Fail on Mobile LiDARs}
Conventional ghost removal methods~\cite{yun2019virtual, gao2022reflective} rely heavily on geometric consistency, where ghost points are detected by comparing them with the corresponding real-object points. This assumption holds for stationary LiDAR systems with a full $360^\circ$ field of view, because both the real object and its ghost reflection are simultaneously observed within the same scan. Consequently, prior approaches can exploit redundant geometric cues to detect and eliminate ghost structures.  
However, this assumption fundamentally breaks down in mobile LiDAR settings. Automotive LiDAR sensors typically provide a narrow field of view of only $80^\circ$–$120^\circ$~\cite{at128, m3}, and the sensor continuously moves through the environment. As a result, the true object behind a glass surface often falls outside the field of view when the ghost is observed. This makes the simultaneous observation of real and ghost points highly unlikely, rendering geometry-based ghost detection intrinsically infeasible for mobile LiDAR. Moreover, the sparsity of mobile LiDAR point clouds and the dynamic nature of real-world scenes further weaken geometric cues, preventing the accumulation of consistent multi-view observations required by prior methods.

\subsection{Evaluation on Downstream Applications}

\subsubsection{SLAM Experimental Scenes}
\cref{fig:slam} presents additional photograph of the SLAM environment together with the corresponding trajectory used in our experiment.
The SLAM sequence was captured in an office corridor where strong glass reflections frequently produce ghost points, providing a challenging setting for evaluating the effectiveness of our ghost removal.

\subsubsection{SLAM Ablation Studies}
We compare against LiDAR signal processing strategies using commercial LiDAR's common factory default settings: Dual-Peak~\cite{ouster,at128} and Multi-Peak~\cite{avia}, which retain the two and three strongest intensity peaks, respectively.
To further examine whether point cloud–based denoising can mitigate ghost artifacts, we conducted supplementary ablation experiments, separate from the main evaluations in this paper.
In these experiments, each peak-selection strategy was combined with one of three preprocessing variants: no filtering (Raw), a Statistical Outlier Filter, or a Radius-based Outlier Filter.
Both filtering modules are provided by Open3D~\cite{zhou2018open3d}.
For the Statistical Outlier Filter, we used a neighbor size of 20 and a standard deviation threshold of 2.0.
For the Radius-based Outlier Filter, we set a minimum point count of 50 within a search radius of 0.5 m.
We also applied Open3D's voxel downsampling with a voxel size of 1.0 for all methods.
All methods employed the same SLAM backend, GLIM~\cite{koide2024glim}, ensuring a fair comparison.

We report the SLAM ablation results in Table~\ref{tab:slam_comparison_transposed}.
Although point-cloud–based denoising methods provide a modest improvement in SLAM accuracy, our waveform-based ghost removal delivers a larger and consistent accuracy gain.
When compared to the baseline methods combined with the Statistical Outlier Filter, our ghost removal further reduces ATE by 54--76\% and RTE by 53--78\%.
Similarly, relative to the Radius-based Outlier Filter, ATE decreases by 35–63\% and RTE by 32–62\%.
These results demonstrate that conventional outlier filtering is insufficient for handling ghost artifacts, and that our waveform-driven ghost removal provides substantially more effective suppression, leading to the highest SLAM accuracy.

\begin{table}[t]
\centering
\caption{Details of the object-detection test dataset (PedScene).}
\label{tab:object-detection-scene}
\setlength{\tabcolsep}{4pt}
\renewcommand{\arraystretch}{0.95}
\begin{tabularx}{0.48\textwidth}{l*{4}{>{\centering\arraybackslash}X}}
\toprule
Scene & \begin{tabular}{@{}c@{}}Indoor/\\Outdoor\end{tabular} & Pedestrians & Frames \\
\midrule
PedScene 001 & Indoor  & 2 & 14 \\
PedScene 002 & Outdoor & 3 & 31 \\
PedScene 003 & Indoor  & 3 & 32 \\
PedScene 004 & Indoor  & 1 & 25 \\
\midrule
\textbf{TOTAL} 
  & \textbf{\begin{tabular}{@{}c@{}}Indoor 3 /\\ Outdoor 1\end{tabular}} 
  & \textbf{1 / 2 / 3} 
  & \textbf{102} \\
\bottomrule
\end{tabularx}
\end{table}

\begin{figure}[t]
  \begin{center}
    \includegraphics[width=1.0\linewidth]{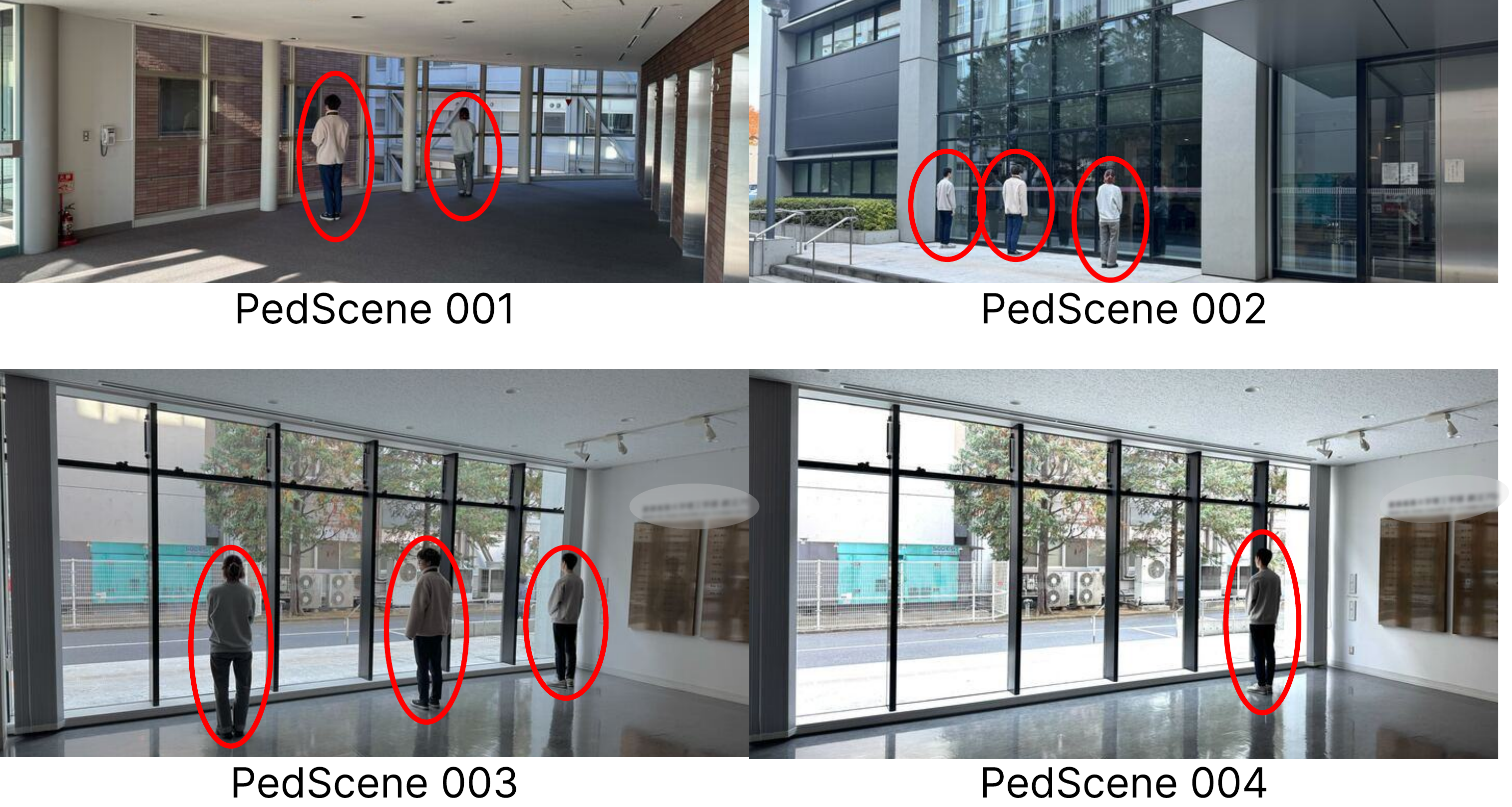}
    \caption{
        Example scenes from the PedScene dataset used for 3D object detection.  
        The dataset includes both indoor and outdoor environments with glass surfaces, building entrances, and glass-walled sidewalks.
    }
    \label{fig:ped_scene}
  \end{center}
  \vspace{-4mm}
\end{figure}

\subsubsection{Object Detection Test Dataset}

We summarize the details of the object-detection test dataset in Table~\ref{tab:object-detection-scene}.  
The dataset, referred to as \textit{PedScene}, consists of four scenes recorded in indoor and outdoor environments containing glass surfaces, building entrances, and glass-walled sidewalks.  
\cref{fig:ped_scene} shows examples of these environments together with RGB images.

{
    \small
    \bibliographystyle{ieeenat_fullname}
    \bibliography{main, supplement}

@inproceedings{sun2020waymo,
  title={{Scalability in perception for autonomous driving: Waymo open dataset}},
  author={Sun, Pei and Kretzschmar, Henrik and Dotiwalla, Xerxes and Chouard, Aurelien and Patnaik, Vijaysai and Tsui, Paul and Guo, James and Zhou, Yin and Chai, Yuning and Caine, Benjamin and others},
  booktitle={IEEE/CVF Conference on Computer Vision and Pattern Recognition (CVPR)},
  pages={2446--2454},
  year={2020}
}

@inproceedings{caesar2020nuscenes,
  title = {{ nuScenes: A Multimodal Dataset for Autonomous Driving }},
  author={Caesar, Holger and Bankiti, Varun and Lang, Alex H and Vora, Sourabh and Liong, Venice Erin and Xu, Qiang and Krishnan, Anush and Pan, Yu and Baldan, Giancarlo and Beijbom, Oscar},
  booktitle={IEEE/CVF Conference on Computer Vision and Pattern Recognition (CVPR)},
  pages={11621--11631},
  year={2020}
}

@article{geiger2013kitti,
  title = {{Vision meets robotics: The KITTI dataset}},
  author={Geiger, Andreas and Lenz, Philip and Stiller, Christoph and Urtasun, Raquel},
  journal={The International Journal of Robotics Research},
  volume={32},
  number={11},
  pages={1231--1237},
  year={2013},
}

@inproceedings{yun2018reflection,
  title={{Reflection Removal for Large-Scale 3D Point Clouds}},
  author={Yun, Jae-Seong and Sim, Jae-Young},
  booktitle={IEEE/CVF Conference on Computer Vision and Pattern Recognition (CVPR)},
  pages={4597--4605},
  year={2018}
}

@article{yun2019virtual,
  title={{ Virtual Point Removal for Large-Scale 3D Point Clouds with Multiple Glass Planes }},
  author={Yun, Jae-Seong and Sim, Jae-Young},
  journal={IEEE Transactions on Pattern Analysis and Machine Intelligence (TPAMI)},
  volume={43},
  number={2},
  pages={729--744},
  year={2019},
}

@article{lee2023learning,
  title={{Learning-Based Reflection-Aware Virtual Point Removal for Large-Scale 3D Point Clouds}},
  author={Lee, Oggyu and Joo, Kyungdon and Sim, Jae-Young},
  journal={IEEE Robotics and Automation Letters},
  volume={8},
  number={12},
  pages={8510--8517},
  year={2023},
}

@inproceedings{zhao2020mapping,
  title={{Mapping with Reflection - Detection and Utilization of Reflection in 3D Lidar Scans}},
  author={Zhao, Xiting and Yang, Zhijie and Schwertfeger, S{\"o}ren},
  booktitle={IEEE International Symposium on Safety, Security, and Rescue Robotics (SSRR)},
  pages={27--33},
  year={2020},
}

@inproceedings{deziel2021pixset,
  title={Pixset: An Opportunity for 3D Computer Vision to Go Beyond Point Clouds with a Full-Waveform LiDAR Dataset},
  author={D{\'e}ziel, Jean--Luc and Merriaux, Pierre and Tremblay, Francis and Lessard, Dave and Plourde, Dominique and Stanguennec, Julien and Goulet, Pierre and Olivier, Pierre},
  booktitle={IEEE International Intelligent Transportation Systems Conference (ITSC)},
  pages={2987--2993},
  year={2021},
}

@article{xu2022fast,
  title={{FAST-LIO2: Fast Direct LiDAR-Inertial Odometry}},
  author={Xu, Wei and Cai, Yixi and He, Dongjiao and Lin, Jiarong and Zhang, Fu},
  journal={IEEE Transactions on Robotics},
  volume={38},
  number={4},
  pages={2053--2073},
  year={2022},
}

@article{koide2024glim,
  title={{GLIM: 3D range-inertial localization and mapping with GPU-accelerated scan matching factors}},
  author={Koide, Kenji and Yokozuka, Masashi and Oishi, Shuji and Banno, Atsuhiko},
  journal={Robotics and Autonomous Systems},
  volume={179},
  pages={104750},
  year={2024},
}

@inproceedings{charron2018noising,
  title={{De-noising of Lidar Point Clouds Corrupted by Snowfall}},
  author={Charron, Nicholas and Phillips, Stephen and Waslander, Steven L},
  booktitle={Conference on Computer and Robot Vision (CRV)},
  pages={254--261},
  year={2018},
}

@inproceedings{he2022mae,
  author={He, Kaiming and Chen, Xinlei and Xie, Saining and Li, Yanghao and Dollár, Piotr and Girshick, Ross},
  booktitle={IEEE/CVF Conference on Computer Vision and Pattern Recognition (CVPR)}, 
  title={{Masked Autoencoders Are Scalable Vision Learners}}, 
  year={2022},
  pages={15979-15988},
}

@inproceedings{tong2022videomae,
  author = {Tong, Zhan and Song, Yibing and Wang, Jue and Wang, Limin},
  title = {{VideoMAE: masked autoencoders are data-efficient learners for self-supervised video pre-training}},
  year = {2022},
  booktitle = {International Conference on Neural Information Processing Systems (NeurIPS)},
  articleno = {732},
  numpages = {16},
}

@inproceedings{pang2022pointmae,
  title={{Masked Autoencoders for Point Cloud Self-supervised Learning}},
  author={Pang, Yatian and Wang, Wenxiao and Tay, Francis EH and Liu, Wei and Tian, Yonghong and Yuan, Li},
  booktitle={European Conference on Computer Vision (ECCV)},
  pages={604--621},
  year={2022},
}

@inproceedings{liu2022maskpoint,
  author = {Liu, Haotian and Cai, Mu and Lee, Yong Jae},
  title = {{Masked Discrimination for Self-supervised Learning on Point Clouds}},
  year = {2022},
  booktitle = {European Conference on Computer Vision (ECCV)},
  pages = {657–675},
}

@INPROCEEDINGS {yu2022pointbert,
  author = { Yu, Xumin and Tang, Lulu and Rao, Yongming and Huang, Tiejun and Zhou, Jie and Lu, Jiwen },
  booktitle = {IEEE/CVF Conference on Computer Vision and Pattern Recognition (CVPR)},
  title = {{Point-BERT: Pre-training 3D Point Cloud Transformers with Masked Point Modeling}},
  year = {2022},
  pages = {19291-19300},
}

@inproceedings {hess2023voxelmae,
  author = { Hess, Georg and Jaxing, Johan and Svensson, Elias and Hagerman, David and Petersson, Christoffer and Svensson, Lennart },
  booktitle = {IEEE/CVF Winter Conference on Applications of Computer Vision Workshops (WACVW) },
  title = {{Masked Autoencoder for Self-Supervised Pre-training on Lidar Point Clouds}},
  year = {2023},
  pages = {350-359},
}

@inproceedings{yang2023gdmae,
  author={Yang, Honghui and He, Tong and Liu, Jiaheng and Chen, Hua and Wu, Boxi and Lin, Binbin and He, Xiaofei and Ouyang, Wanli},
  booktitle={IEEE/CVF Conference on Computer Vision and Pattern Recognition (CVPR)}, 
  title={{GD-MAE: Generative Decoder for MAE Pre-Training on LiDAR Point Clouds}}, 
  year={2023},
  pages={9403-9414},
}

@inproceedings{tian2023geomae,
  author={Tian, Xiaoyu and Ran, Haoxi and Wang, Yue and Zhao, Hang},
  booktitle={IEEE/CVF Conference on Computer Vision and Pattern Recognition (CVPR)}, 
  title={{GeoMAE: Masked Geometric Target Prediction for Self-supervised Point Cloud Pre-Training}}, 
  year={2023},
  pages={13570-13580},
}

@inproceedings{xu2023mvjar,
  author = { Xu, Runsen and Wang, Tai and Zhang, Wenwei and Chen, Runjian and Cao, Jinkun and Pang, Jiangmiao and Lin, Dahua },
  booktitle = {IEEE/CVF Conference on Computer Vision and Pattern Recognition (CVPR)},
  title = {{MV-JAR: Masked Voxel Jigsaw and Reconstruction for LiDAR-Based Self-Supervised Pre-Training}},
  year = {2023},
  pages = {13445-13454},
}

@article{min2024occupancymae,
  author={Min, Chen and Xiao, Liang and Zhao, Dawei and Nie, Yiming and Dai, Bin},
  journal={IEEE Transactions on Intelligent Vehicles}, 
  title={{Occupancy-MAE: Self-Supervised Pre-Training Large-Scale LiDAR Point Clouds With Masked Occupancy Autoencoders}}, 
  year={2024},
  volume={9},
  number={7},
  pages={5150-5162},
}

@article{shen2025marmot,
  title={{MARMOT: Masked Autoencoder for Modeling Transient Imaging}}, 
  author={Siyuan Shen and Ziheng Wang and Xingyue Peng and Suan Xia and Ruiqian Li and Shiying Li and Jingyi Yu},
  year={2025},
  journal={arXiv preprint arXiv:2506.08470},
}

@inproceedings{chen2020simclr,
author = {Chen, Ting and Kornblith, Simon and Norouzi, Mohammad and Hinton, Geoffrey},
title = {{A simple framework for contrastive learning of visual representations}},
year = {2020},
booktitle = {International Conference on Machine Learning (ICML)},
articleno = {149},
numpages = {11},
}

@INPROCEEDINGS {he2020moco,
author = { He, Kaiming and Fan, Haoqi and Wu, Yuxin and Xie, Saining and Girshick, Ross },
booktitle = {IEEE/CVF Conference on Computer Vision and Pattern Recognition (CVPR)},
title = {{Momentum Contrast for Unsupervised Visual Representation Learning}},
year = {2020},
pages = {9726-9735},
}

@article{chen2020moco2,
  title={{Improved Baselines with Momentum Contrastive Learning}}, 
  author={Xinlei Chen and Haoqi Fan and Ross Girshick and Kaiming He},
  year={2020},
  journal={arXiv:2003.04297},
}

@INPROCEEDINGS {chen2021moco3,
author = { Chen, Xinlei and Xie, Saining and He, Kaiming },
booktitle = {IEEE/CVF International Conference on Computer Vision (ICCV)},
title = {{An Empirical Study of Training Self-Supervised Vision Transformers}},
year = {2021},
pages = {9620-9629},
}

@INPROCEEDINGS {lin2017focal,
author = { Lin, Tsung-Yi and Goyal, Priya and Girshick, Ross and He, Kaiming and Dollar, Piotr },
booktitle = {IEEE International Conference on Computer Vision (ICCV)},
title = {{Focal Loss for Dense Object Detection}},
year = {2017},
pages = {2999-3007},
}

@inproceedings{scheublelidar,
author    = {Scheuble, Dominik and Holzh\"uter, Hanno and Peters, Steven and Bijelic, Mario and Heide, Felix},
    title     = {{Lidar Waveforms are Worth 40x128x33 Words}},
    booktitle = {IEEE/CVF International Conference on Computer Vision (ICCV)},
    year      = {2025},
    pages     = {28913-28924}
}

@article{gao2022reflective,
  title={{Reflective Noise Filtering of Large-Scale Point Cloud Using Transformer}},
  author={Gao, Rui and Li, Mengyu and Yang, Seung-Jun and Cho, Kyungeun},
  journal={Remote Sensing},
  volume={14},
  number={3},
  pages={577},
  year={2022},
}

@article{gaydon2024fractal,
  title={{FRACTAL: An Ultra-Large-Scale Aerial Lidar Dataset for 3D Semantic Segmentation of Diverse Landscapes}},
  author={Gaydon, Charles and Daab, Michel and Roche, Floryne},
  journal={arXiv preprint arXiv:2405.04634},
  year={2024}
}

@inproceedings{dosovitskiy2017carla,
  title = {{{CARLA}: {An} Open Urban Driving Simulator}},
  author = {Alexey Dosovitskiy and German Ros and Felipe Codevilla and Antonio Lopez and Vladlen Koltun},
  booktitle = {Conference on Robot Learning (CoRL)},
  pages = {1--16},
  year = {2017}
}

@misc{Mitsuba3,
    title = {{Mitsuba 3 renderer}},
    author = {Wenzel Jakob and Sébastien Speierer and Nicolas Roussel and Merlin Nimier-David and Delio Vicini and Tizian Zeltner and Baptiste Nicolet and Miguel Crespo and Vincent Leroy and Ziyi Zhang},
    note = {https://mitsuba-renderer.org},
    version = {3.1.1},
    year = 2022
}

@misc{autoware,
  title = {{Home Page}},
  author = {Autoware Foundation},
  howpublished = {https://autoware.org},
  year = {2025}
}

@misc{airborne-lidar,
  title = {{Airborne Lidar Products}},
  author={Teledyne Technologies Inc},
  howpublished = {https://www.teledyneoptech.com/products/airborne-lidar},
  year = {2025}
}

@misc{apollo,
  title = {{Intelligent Vehicle}},
  author = {{apollo}},
  howpublished = {https://www.apollo.auto/apollo-self-driving},
  year = {2025}
}

@misc{mid-360,
    title={{Mid-360}},
    author={{Livox}},
    howpublished = {https://www.livoxtech.com/mid-360},
    year = {2025}
}

@misc{unitreeg1,
  title = {{Unitree G1}},
  author={{Unitree Robotics}},
  howpublished={https://www.unitree.com/g1},
  year = {2025}
}

@misc{spot,
 title= {{Spot - The Agile Mobile Robot}},
 author={{Boston Dynamics.}},
 howpublished={https://bostondynamics.com/products/spot/},
  year = {2025}
}

@misc{at128,
  title = {{AT128 Automotive-Grade 120° Long-Range Lidar - Hesai}},
  author={{Hesai}},
  howpublished={https://www.hesaitech.com/product/at128},
  year = {2025}
}

@misc{m3,
  title = {{Robosense M3}},
  author = {{RoboSense Technology Co., Ltd.}},
  howpublished={https://www.robosense.ai/en/rslidar/M3},
  year = {2025}
}

@misc{avia,
  title = {{Avia}},
  author = {{Livox}},
  howpublished = {https://www.livoxtech.com/avia},
  year = {2025}
}

@misc{ouster,
  title = {{OS1 Mid-Range High-Resolution Imaging Lidar}},
  author= {{Ouster Inc}},
  howpublished={https://data.ouster.io/downloads/datasheets/datasheet-rev7-v3p1-os1.pdf},
  year = {2025}
}

@inproceedings{Shi_PV-RCNN_Point-Voxel_Feature_Set_Abstraction_CVPR_2020,
    author = {Shi, Shaoshuai and Guo, Chaoxu and Jiang, Li and Wang, Zhe and Shi, Jianping and Wang, Xiaogang and Li, Hongsheng},
    title = {{PV-RCNN: Point-Voxel Feature Set Abstraction for 3D Object Detection}},
    booktitle = {IEEE/CVF Conference on Computer Vision and Pattern Recognition (CVPR)},
    pages = {10529--10538},
    year = {2020}
}

@inproceedings {sato-dirty,
author = {Takami Sato and Junjie Shen and Ningfei Wang and Yunhan Jia and Xue Lin and Qi Alfred Chen},
title = {{Dirty Road Can Attack: Security of Deep Learning based Automated Lane Centering under {Physical-World} Attack}},
booktitle = {30th USENIX Security Symposium (USENIX Security 21)},
year = {2021},
pages = {3309--3326},
}

@inproceedings{tanabe2019inter,
  title={{Inter-Frame Smart-Accumulation Technique for Long-Range and High-Pixel Resolution LiDAR}},
  author={Tanabe, Ken and Kubota, Hiroshi and Sai, Akihide and Matsumoto, Nobu},
  booktitle={IEEE Symposium in Low-Power and High-Speed Chips (COOL CHIPS)},
  pages={1--3},
  year={2019},
}

@article{yoshioka201820,
  title={{A 20-ch TDC/ADC Hybrid Architecture LiDAR SoC for 240x96 Pixel 200-m Range Imaging With Smart Accumulation Technique and Residue Quantizing SAR ADC}},
  author={Yoshioka, Kentaro and Kubota, Hiroshi and Fukushima, Tomonori and Kondo, Satoshi and Ta, Tuan Thanh and Okuni, Hidenori and Watanabe, Kaori and Hirono, Masatoshi and Ojima, Yoshinari and Kimura, Katsuyuki and others},
  journal={IEEE Journal of Solid-State Circuits},
  volume={53},
  number={11},
  pages={3026--3038},
  year={2018},
}

@inproceedings{reynolds2011capturing,
  title={{Capturing time-of-flight data with confidence}},
  author={Reynolds, Malcolm and Dobo{\v{s}}, Jozef and Peel, Leto and Weyrich, Tim and Brostow, Gabriel J},
  booktitle={IEEE/CVF Conference on Computer Vision and Pattern Recognition (CVPR)},
  pages={945--952},
  year={2011},
}

@article{mallet2009full,
  title={{Full-waveform topographic lidar: State-of-the-art}},
  author={Mallet, Cl{\'e}ment and Bretar, Fr{\'e}d{\'e}ric},
  journal={ISPRS Journal of photogrammetry and remote sensing},
  volume={64},
  number={1},
  pages={1--16},
  year={2009},
}

@article{zou2024256,
  title={{A 256x192-Pixel Direct Time-of-Flight LiDAR Receiver With a Current-Integrating-Based AFE Supporting 240-m-Range Imaging}},
  author={Zou, Chaorui and Ou, Yaozhong and Zhu, Yan and Martins, Rui P and Chan, Chi-Hang and Zhang, Minglei},
  journal={IEEE Journal of Solid-State Circuits},
  year={2024},
}

@article{lindell2018single,
author = {Lindell, David B. and O'Toole, Matthew and Wetzstein, Gordon},
title = {{Single-photon 3D imaging with deep sensor fusion}},
year = {2018},
volume = {37},
number = {4},
articleno = {113},
numpages = {12},
journal = {ACM Trans. Graph.},
}

@inproceedings{zhang2018tutorial,
  title={{A Tutorial on Quantitative Trajectory Evaluation for Visual(-Inertial) Odometry}},
  author={Zhang, Zichao and Scaramuzza, Davide},
  booktitle={IEEE/RSJ international conference on intelligent robots and systems (IROS)},
  pages={7244--7251},
  year={2018},
}

@inproceedings{kraus2021radar,
  title={{The Radar Ghost Dataset – An Evaluation of Ghost Objects in Automotive Radar Data}},
  author={Kraus, Florian and Scheiner, Nicolas and Ritter, Werner and Dietmayer, Klaus},
  booktitle={IEEE/RSJ International Conference on Intelligent Robots and Systems (IROS)},
  pages={8570--8577},
  year={2021},
}

@InProceedings{cicek2016unet3d,
author="{\c{C}}i{\c{c}}ek, {\"O}zg{\"u}n
and Abdulkadir, Ahmed
and Lienkamp, Soeren S.
and Brox, Thomas
and Ronneberger, Olaf",
editor="Ourselin, Sebastien
and Joskowicz, Leo
and Sabuncu, Mert R.
and Unal, Gozde
and Wells, William",
title={{3D U-Net: Learning Dense Volumetric Segmentation from Sparse Annotation}},
booktitle="Medical Image Computing and Computer-Assisted Intervention -- MICCAI 2016",
year="2016",
pages="424--432",
}

@inproceedings{Loshchilov2017DecoupledWD,
  title={{Decoupled Weight Decay Regularization}},
  author={Ilya Loshchilov and Frank Hutter},
  booktitle={International Conference on Learning Representations (ICLR)},
  year={2017},
}

@article{Sager_2022,
    doi = {10.14733/cadaps.2022.1191-1206},
    url = {http://cad-journal.net/files/vol_19/CAD_19(6)_2022_1191-1206.pdf},
    year = 2022,
    month = {mar},
    publisher = {{CAD} Solutions, {LLC}},
    volume = {19},
    number = {6},
    pages = {1191--1206},
    author = {Christoph Sager and Patrick Zschech and Niklas Kuhl},
    title = {{labelCloud}: A Lightweight Labeling Tool for Domain-Agnostic 3D Object Detection in Point Clouds},
    journal = {Computer-Aided Design and Applications}
}

@article{zhou2018open3d,
  title={{Open3D: A modern library for 3D data processing}},
  author={Zhou, Qian-Yi and Park, Jaesik and Koltun, Vladlen},
  journal={arXiv preprint arXiv:1801.09847},
  year={2018}
}

@String(CVPR= {IEEE Conf. Comput. Vis. Pattern Recog.})

@String(ICCV= {Int. Conf. Comput. Vis.})

@String(ECCV= {Eur. Conf. Comput. Vis.})

@String(ICLR = {Int. Conf. Learn. Represent.})

@String(CVPR  = {CVPR})

@String(ICCV  = {ICCV})

@String(ECCV  = {ECCV})

@String(ICLR  = {ICLR})
}

\end{document}